\documentclass[final]{alt2026} % Include author names

\title[Universal Dynamic Regret and Violation Bounds for Constrained OCO]{Universal Dynamic Regret and Constraint Violation Bounds for Constrained Online Convex Optimization}
\usepackage{times}
\altauthor{%
 \Name{Subhamon Supantha}$^{*}$ \Email{subhamon@minds.iitb.ac.in}\\
 \addr Indian Institute of Technology, Bombay, India
 \AND
 \Name{Abhishek Sinha} \Email{abhishek.sinha@tifr.res.in}\\
 \addr Tata Institute of Fundamental Research, Mumbai, India%
}
\usepackage{booktabs}       % professional-quality tables
\usepackage{color}
\usepackage[utf8]{inputenc} % allow utf-8 input
\usepackage[T1]{fontenc}    % use 8-bit T1 fonts
\usepackage{hyperref}       % hyperlinks
\usepackage{url}            % simple URL typesetting
\usepackage{booktabs}       % professional-quality tables
\usepackage{amsfonts}       % blackboard math symbols
\usepackage{nicefrac}       % compact symbols for 1/2, etc.
\usepackage{microtype}      % microtypography
\usepackage{setspace}
\usepackage{dsfont}
\usepackage{amsmath}
\usepackage{amssymb, bm}
\usepackage[dvipsnames,svgnames]{xcolor}
\usepackage{xcolor}
\usepackage{hyperref}
\usepackage{mathrsfs}
\usepackage{mathtools}
\usepackage{algorithm}
\usepackage{algorithmic}
\usepackage{graphicx}

\setcitestyle{authoryear,open={(},close={)}} %Citation-related commands

\newcounter{commentcounter}
\newcommand{\cmt}[1]{%
  \refstepcounter{commentcounter}%
  \textcolor{red}{\textbf{(\thecommentcounter) AS: #1}}%
}

\newtheorem{assumption}{Assumption}

\hypersetup{
	colorlinks=true,
	linkcolor=blue,
	urlcolor=red,
	citecolor=NavyBlue,
	linkbordercolor={0 0 1}
}
\newcommand{\patht}[1]{\mathcal{P}_t(#1)}

\newcommand{\Xstart}{\mathcal{X}^*_t}
\newcommand{\dregt}[1]{\textrm{D-Reg}_t(#1)}

\newcommand{\dregdasht}[1]{\textrm{D-Reg'}_t(#1)}

\newcommand{\ftilde}{\tilde{f}}

\newcommand{\fhat}{\hat{f}}
\newcommand{\sumlim}{\sum\limits}
\newcommand{\dist}{\mathsf{dist}}

\begin{document}
\maketitle
\begingroup
\makeatletter
\renewcommand\@makefntext[1]{#1}
\footnotetext{${}^{*}$This work was initiated while the first author was working as a summer intern at TIFR, Mumbai.}
\makeatother
\endgroup
% If your paper is accepted and the title of your paper is very long,
% the style will print as headings an error message. Use the following
% command to supply a shorter title of your paper so that it can be
% used as headings.
%
%\runningtitle{I use this title instead because the last one was very long}

% If your paper is accepted and the number of authors is large, the
% style will print as headings an error message. Use the following
% command to supply a shorter version of the authors names so that
% they can be used as headings (for example, use only the surnames)
%
%\runningauthor{Surname 1, Surname 2, Surname 3, ...., Surname n}

%\twocolumn[
%
%\aistatstitle{Universal Dynamic Regret Bounds for Learning with Adversarial Constraints}
%
%\aistatsauthor{ Subhamon Supantha \And Abhishek Sinha}
%
%\aistatsaddress{  CMI \And TIFR, Mumbai } 
%]

\begin{abstract}
We consider a generalization of the celebrated Online Convex Optimization (OCO) framework with adversarial online constraints. In this problem, an online learner interacts with an adversary sequentially over multiple rounds. At the beginning of each round, the learner chooses an action from a convex decision set. After that, the adversary reveals a convex cost function and a convex constraint function. The goal of the learner is to minimize the cumulative cost while satisfying the constraints as tightly as possible. We present two efficient algorithms with simple modular structures that give universal dynamic regret and cumulative constraint violation bounds, improving upon state-of-the-art results. While the first algorithm, which achieves the optimal regret bound, involves projection onto the constraint sets, the second algorithm is projection-free and achieves better violation bounds in rapidly varying environments. Our results hold in the most general case when both the cost and constraint functions are chosen arbitrarily, and the constraint functions need not contain any fixed common feasible point. We establish these results by introducing a general framework that reduces the constrained learning problem to an instance of the standard OCO problem with specially constructed surrogate cost functions. 
\end{abstract}

\section{Introduction} \label{intro}
%\cmt{Resize the table}
Online Convex Optimization (OCO) is a standard framework for studying sequential decision-making under uncertainty. In this framework, an online learner selects an action $x_t$ from a convex decision set $\mathcal{X} \subseteq \mathbb{R}^d,$ for $T$ rounds ($t=1,2, \ldots, T$). At the end of each round $t,$ an adversary reveals a convex cost function $f_t: \mathcal{X} \mapsto \mathbb{R}.$ Consequently, the learner incurs a cost of $f_t(x_t)$ for round $t.$ The goal of the learner is to choose actions sequentially to minimize the cumulative cost $\sum_{t=1}^T f_t(x_t)$ over the horizon of length $T$ \citep{hazan2022introduction, orabona2019modern}.
% Online Learning problem where the aim is to design a policy to choose at each time step $t$, a move $x_t$ from some convex set, incurring a cost of $f_t(x_t)$, where $f_t$ is a convex loss function chosen(possibly adversarially) after $x_t$ is played. The aim of the algorithm is to minimize the difference between the cumulative losses of its choices and a single best action in hindsight, or \emph{regret}. \citep{HazanBook, orabona2019modern}. 
  Since \citet{zinkevich2003online} introduced the ubiquitous Online Gradient Descent (OGD) algorithm and showed that it achieves the minimax optimal $O(\sqrt{T})$ static regret, many improved algorithms have been proposed in the literature with refined analyses and sharper performance guarantees. See \citet{hazan2022introduction} and \citet{orabona2019modern} for excellent textbook treatments of this topic.  
  %They also showed that this is the minimax optimal rate. 
%In other words, they showed that the problem, in the generality they were considering, could not hope to achieve better than the $\mathcal{O}({\sqrt{T}})$ regret due to the presence of an adversarial strategy which can force the algorithm to incur $\Omega(\sqrt{T})$ cost.

%Since then, different variations, settings and generalizations of the OCO problem have been studied. A detailed survey of the literature is provided in \citet{HazanBook} and \citet{orabona2019modern}.

%A substantially harder metric to determine the performance of a Online Learning algorithm is the \emph{dynamic regret}, where the actions in hindsight against which the cumulative loss of the actions of the learner is benchmarked is allowed to be time-varying and not just a fixed best action.

%It is customary in this case to bound the regret in terms of the path length $\mathcal{P}_T$, a measure of variability, of the benchmarking sequence.  

A substantial generalization of OCO called Constrained Online Convex Optimization (COCO), which involves additional time-varying constraints, has recently attracted sufficient attention from the learning community \citep{mahdavi2012trading, cao2018online, neely2017online,guo2022online,sinha2024optimal,lekeufack2024optimistic}. COCO arises in many settings, including AI safety \citep{safeRL, hua2025combining}, obstacle avoidance in robotics \citep{snyder2023online}, Bandits with Knapsacks \citep{immorlica2022adversarial}, and pay-per-click ad markets with budget constraints \citep{georgios-cautious}. In this problem, after the online learner takes its action $x_t \in \mathcal{X}$ at round $t$, in addition to the cost function $f_t,$ a constraint function $g_t : \mathcal{X} \mapsto \mathbb{R}$ is revealed to the learner. The function $g_t$ encodes an instantaneous constraint of the form $g_t(x) \leq 0.$ The online learner is ideally expected to have played an action $x_t$ such that it satisfies the constraint \emph{i.e.,} $g_t(x_t) \leq 0, ~\forall t$. However, this is not always possible as $g_t$ is adversarially chosen and revealed only \emph{after} the learner chooses its action $x_t$. The learner is thus allowed to violate the constraints to some extent, and the goal is to design an online algorithm that simultaneously minimizes the regret for the cost functions and the cumulative constraint violations (CCV) for the constraint functions.
 
To the best of our knowledge, all previous works on the COCO problem, including \citet{sinha2024optimal, guo2022online, cao2018online, neely2017online}, make the following strong assumption on the choice of the constraint functions:
\begin{assumption}[Common Feasibility]  \label{com-feas}
	There exists a fixed action $x^\star \in \mathcal{X}$ which satisfies every constraints, \emph{i.e.,} $g_t(x^\star) \leq 0, \forall t \geq 1. $
\end{assumption}
The common feasibility assumption is substantially restrictive as it limits the flexibility of the adversary in choosing the constraint functions. In practice, if the environment is rapidly varying with time, this assumption may no longer hold and hence, the theoretical guarantees derived under this assumption become vacuous. Dropping Assumption \ref{com-feas} fundamentally changes the nature of the problem. In particular, in the absence of a single action that satisfies all constraints simultaneously, static regret is no longer a meaningful performance metric.
%Under Assumption \ref{com-feas}, \cite{sinha2024optimal} recently proposed a Lyapunov-function based algorithm that yields the minimax optimal static regret and CCV bounds (up to logarithmic factors). 
 
In this paper, we drop Assumption \ref{com-feas} altogether and focus on the stronger objective of guaranteeing universal dynamic regret bounds against \emph{any} feasible comparator sequence. Dynamic regret generalizes the classic static regret metric. While static regret compares the performance of an online algorithm with a \emph{fixed} benchmark, in dynamic regret, we measure the online algorithm's performance against a class of time-varying benchmark sequences.    
Two different variants of dynamic regret metrics are widely studied in the literature - the more conservative \emph{worst-case} dynamic regret and the more flexible \emph{universal} dynamic regret. Guaranteeing universal dynamic regret bounds is technically more challenging as, unlike worst-case dynamic regret, where the path-length of the comparator sequence can be estimated (\emph{e.g.,} using the doubling trick), minimizing universal dynamic regret requires designing algorithms which are oblivious to the comparator path-length. See Section \ref{statement} for details.

In this paper, we propose a general framework, given by Algorithm \ref{COCOAlgogen}, that reduces the constrained online optimization problem to the standard OCO problem. COCO algorithms constructed within this framework only differ in the way the surrogate costs are constructed and the choice of the OCO subroutines used to minimize the surrogate cost functions. 
\begin{algorithm}
\caption{Algorithmic Template for COCO} \label{COCOAlgogen}
\begin{algorithmic}[1]
%    \STATE \textbf{Parameters}: $V=1,\Phi(x) = \exp(\lambda x) - 1,c=2G(D+1),\lambda = \frac{1}{2c \sqrt{1 + \mathcal{P}_T^\star} \sqrt{T}}$   
 %   \STATE \textbf{Initialization}: Set $x_1 = \mathbf{0}~, Q(0)=0$
 \REQUIRE  A generic OCO subroutine $\mathcal{A}^{\textrm{OCO}}$ with a universal dynamic regret bound
 \FOR {$t=1:T$}
 \STATE From the first $t-1$ cost and constraint functions, construct a surrogate cost function $\hat{f}_{t-1}.$  
 \STATE Choose action $x_t$ by running $\mathcal{A}^{\textrm{OCO}}$ on the surrogate cost functions $\{\hat{f}_\tau\}_{\tau=1}^{t-1}$.
 \STATE Receive new cost function $f_t$ and the constraint function $g_t.$
 \ENDFOR
\end{algorithmic}
\end{algorithm}

We emphasize the online nature of Algorithm \ref{COCOAlgogen}: the surrogate cost function $\hat{f}_{t-1}$ and the action $x_t$ depend only on the past $t-1$ cost and constraint functions $\{f_\tau, g_\tau\}_{\tau=1}^{t-1}.$ The cost function $f_t$ and the constraint function $g_t$ are revealed only after $x_t$ is played. Consequently, the cost $f_t$, the constraint $g_t,$ and hence, the surrogate cost $\hat{f}_t$ may depend on the learner's action $x_t.$

\paragraph{Our Contribution}
\begin{enumerate}
\item We present two online algorithms for COCO with different performance and computational tradeoffs. While the first algorithm guarantees a minimax optimal universal dynamic regret bound, the second algorithm achieves a better CCV bound in rapidly varying environments. Furthermore, the first algorithm naturally generalizes to \emph{quasiconvex} constraint functions, beyond the convex setting considered in most prior work. 
\item From a computational viewpoint, while the first algorithm requires the full knowledge of the constraint functions and involves a projection step, the second algorithm is lightweight and requires only first-order gradient information.
%We emphasize that unlike the worst-case dynamic regret metric, which holds for one specific choice of the comparator benchmarks, a universal dynamic regret bound is much stronger as it holds for all feasible comparator sequences. 
%However, proving universal dynamic regret bounds are technically challenging, as unlike its worst-case counterpart, the path length of the comparator sequence is not known \emph{a priori}. 

\item Unlike prior works, our results hold without the restrictive common feasibility assumption (Assumption \ref{com-feas}). Both algorithms follow a conceptually simple template of running a standard OCO subroutine on a sequence of \emph{surrogate} cost functions, as outlined in Algorithm \ref{COCOAlgogen}.
    \item As an intermediate technical result, in Section \ref{prelims}, we present a new Lipschitz-adaptive algorithm, called \textsf{AHAG}, for the standard OCO problem. This algorithm achieves $\tilde{\mathcal{O}}(\sqrt{(1 + \mathcal{P}_T)T})$ universal dynamic regret without \emph{a priori} knowledge of the common Lipschitz constant of the cost functions. This result might be of independent interest.
%    \item For the COCO problem, we extend the work of \cite{sinha2024optimal} to show that their algorithm, with parameters tuned a bit carefully, achieves $\tilde{\mathcal{O}}(\sqrt{(1 + \mathcal{P}_T)T})$ worst-case dynamic regret and CCV with no further assumptions.
%    \item Under much more relaxed assumptions about the feasibility and knowledge of the path length, we give a novel algorithm which achieves $\tilde{\mathcal{O}}((1 + \mathcal{P}_T)\sqrt{T})$ regret and $\tilde{\mathcal{O}}(\sqrt{(1 + \mathbb{P}_T)}T^{\frac{3}{4}})$ CCV in the COCO problem. Here $\mathbb{P}_T$ is the minimum possible path length for any feasible comparator sequence.  
\end{enumerate}

%Another unnatural assumption is the knowledge of the path length $\mathcal{P}_T$ beforehand. This is not an acceptable assumption because the comparator sequence is chosen after the game is over and can be chosen arbitrarily. The assumption of the path length being known limits what comparators can be used to compare the performance of the learner. That is why we do not assume such knowledge beforehand and our algorithm would work for any valid path length. 

\subsection{Related Works}
\paragraph{COCO with Static Regret Guarantees:} Static regret metric in the context of COCO uses a fixed benchmark which is assumed to be feasible for all constraints (see Assumption \ref{com-feas}). Early works on COCO focused on the setting where the constraint functions are fixed throughout. The main motivation behind this line of work is to design computationally efficient first-order algorithms that avoid projecting onto the fixed constraint set \citep{mahdavi2012trading,jenatton2016adaptive,yu2020low, yuan2018online}. \citet{mannor2009online,  yu2017online, sinha2023banditq} considered the problem with stochastic constraints in both full and bandit feedback settings. The more difficult problem of COCO with time-varying adversarial constraints was considered by \cite{neely2017online,yi2022regret}. These papers additionally assume \emph{Slater's condition}, which is more restrictive than common feasibility. Slater's condition assumes that all constraints are strictly satisfied by a positive margin $\eta >0$  by a fixed benchmark. Using only the common feasibility assumption, \citet{guo2022online} proposed an online primal-dual algorithm that achieves $O(\sqrt{T})$ static regret and $O(T^{\nicefrac{3}{4}})$ CCV. More recently, \citet{sinha2024optimal} proposed the first minimax-optimal COCO algorithm that yields $O(\sqrt{T})$ static regret and $\tilde{O}(\sqrt{T})$ CCV. Their algorithm uses a Lipschitz-adaptive variant of the online gradient descent (OGD) subroutine on a sequence of surrogate cost functions which are constructed by linearly combining the cost and constraint functions via an exponential Lyapunov potential. \citet{lekeufack2024optimistic} extended their algorithm to the optimistic setting where the learner has access to unreliable predictions for the cost and constraint functions. Further works have provided more refined CCV guarantees \citep{sinha2025beyond, vaze2025osqrtt}. 
\paragraph{Dynamic Regret bounds in OCO:} In the dynamic regret metric, the performance of an online policy is compared against a sequence of time-varying benchmarks.
The seminal work of \cite{zinkevich2003online} established that the standard OGD policy also achieves $\mathcal{O}((1+\mathcal{P}_T)\sqrt{T})$ universal dynamic regret for the standard OCO problem, where $\mathcal{P}_T$ is the path-length of the benchmark sequence. However, this fell short of the $\Omega(\sqrt{(1+\mathcal{P}_T)T})$ lower bound established by \cite{zhang2018adaptive}, who further proposed a novel algorithm, called \textsf{ADER}, which matched the minimax lower bound. Their algorithm employed the \textsf{Hedge} policy \citep{cesa2006prediction} to track experts, each running an instance of OGD with their own guesses of the path length. The \textsf{ADER} algorithm, however, needs the knowledge of the upper bound of the $\ell_\infty$ norms of the loss functions. \cite{zhao2020dynamic,zhao2024adaptivity} subsequently improved this bound by devising an ensemble learning algorithm which adapts to the smoothness and gradient variation of the loss functions. \cite{yang2016tracking} showed that OGD achieves $\mathcal{O}(\sqrt{(1+\mathcal{P}^\star_T)T})$ worst-case dynamic regret in the general case, where $\mathcal{P}^\star_T$ is the path-length corresponding to the minimizers of the cost functions. Under additional structural assumptions, tighter worst-case dynamic regret bounds have now been established \citep{yang2016tracking,mokhtari2016online,zhang2017improved,besbes2015non, zhang2022simple,chen2017non}.

%A common limitation of this line of works is that they assume that there exists a single action $x^\star$ that satisfies all constraints, \emph{i.e.,} $g_t(x^\star) \leq 0, \forall t\in [T]$. The static regret is computed with respect to this benchmark. However, this greatly limit the applicability of these algorithms in the fully adversarial setting where the feasible sets may have empty intersection. 

%Recently, \cite{lekeufack2024optimistic} gave an optimistic online mirror descent based approach which achieved better problem dependent bound while matching \cite{sinha2024optimal} in the worst case. A lot of these works have looked at the case of soft constraints - where the learner tries to minimize $\sum\limits_{t=1}^Tg_t(x_t)$ instead of $\sum\limits_{t=1}^Tg_t^+(x_t)$. 

%\paragraph{Dynamic COCO}
\paragraph{COCO with Dynamic Regret Guarantees:} Coming back to the COCO problem, for time-invariant constraint functions, \cite{yi2021regret} achieved $\mathcal{O}(\sqrt{(1+\mathcal{P}_T^\star)T})$ worst-case dynamic regret and $O(\sqrt{T})$ CCV, which was later improved to $O(\sqrt{(1+\mathcal{P}_T^\star)T})$ regret and $O(\log{T})$ CCV by \cite{guo2022online}. In case of time-varying adversarial constraints, \cite{guo2022online} achieved $\mathcal{O}(\sqrt{(1+\mathcal{P}_T)T})$ universal dynamic regret and $O(T^{\frac{3}{4}})$ CCV under the Common Feasibility assumption (Assumption \ref{com-feas}). Under both Assumption \ref{com-feas} and known path length, \cite{lekeufack2024optimistic} proposed an algorithm with worst case dynamic regret $O(\sqrt{(1+\mathcal{P}_T)T})$ and $\tilde{O}(\sqrt{T})$ CCV guarantees. \cite{liu2021simultaneously} proposed an algorithm with a worst-case dynamic regret bound of $O(\sqrt{(1+\mathcal{P}_T^\star)T})$ and $O(\max(\sqrt{T},V_g))$ CCV, where $V_g \equiv \sum\limits_{t=1}^{T} \sup_{x \in \mathcal{X}}  ||g_t(x) - g_{t-1}(x)||_2$. A major drawback of this result is that $V_g$ can be linear in $T$ even for very simple problem instances (\emph{e.g.,} when the constraints are chosen alternatively from two functions $g_1, g_2$ where $g_1\neq g_2.$)
%So in the case where the function variation is $O(T)$, their algorithm also suffers the same order of CCV. The constraint type they consider is also soft.
Table \ref{comp-table} compares and contrasts the results from the previous works with respect to ours.
\begin{table*}[!h]\label{Table:Related Works}
\hspace*{-0.9cm}
%\vskip 0.15in
%\begin{center}
\centering
%\begin{small}
%\begin{sc}
\scalebox{.8}{
\begin{tabular}{llllll} %lccccr
\toprule
Reference & Regret & Type & CCV & Complexity & Assumptions\\
\midrule
% \citet{mahdavi2012trading} & $O(\sqrt{T})$& $O(T^{\nicefrac{3}{4}})$& Fixed Constraints \\\citet{jenatton2016adaptive}& $O(T^{\max(\beta, 1-\beta) })$& $O(T^{1- \beta/2})$& Fixed Constraints& \\
% \citet{yu2017online}    & $O(\sqrt{T})$& $O(\sqrt{T})$& Slater \& Stochastic Constr. &       \\
% \citet{neely2017online}     & $O(\sqrt{T})$ & $O(\sqrt{T})$& Slater&\\
% \citet{yu2020low} & $O(\sqrt{T})$ & $O(1)$& Slater \& Fixed Constraints&\\
% \citet{yi2021regret}     & $O(\sqrt{T})$& $O(T^{\nicefrac{1}{4}})$& Fixed Constraints&\\
% \citet{yi2023distributed}    & $O(T^{\max(\beta,1-\beta)})$& $O(T^{1-\beta})$& Slater &\\   
%\citet{sinha2023playing}    & $O(\sqrt{T})$& $O(T^{\nicefrac{3}{4}})$& Proj& ---&        \\
%\citet{sinha2023playing}    & $O(\sqrt{T})$& $O(\sqrt{T})$& Proj& $\textrm{Regret}_T \geq 0$ &       \\
% \citet{garber2024projectionfree}   & $O(T^{\nicefrac{3}{4}})$& $O(T^{\nicefrac{7}{8}})$& --- &\\
% \citet{garber2024projectionfree}   & $O(T^{\nicefrac{3}{4}})$& $O(T^{\nicefrac{7}{8}})$& --- &\\
% \citet{sinha2024optimal}   & $O(\sqrt{T})$& $\tilde{O}(\sqrt{T})$& --- &\\
\citet{yi2021regret}    &$\mathcal{O}(\sqrt{(1+\mathcal{P}_T(u_{1:T}))T})$&\textbf{UD} & $\mathcal{O}({\sqrt{T}})$ & \textsc{Conv-OPT} & Fixed Constr.\\
%\citet{guo2022online}    &$\mathcal{O}(\sqrt{(1+\mathcal{P}_T)T})$\textbf{(UD)} & $\mathcal{O}({\log{T}})$& Fixed Constraint& \textsc{Conv-OPT} & ---\\
\citet{guo2022online}    & $\mathcal{O}((1+\mathcal{P}_T(u_{1:T}))\sqrt{T})$&\textbf{UD} & $\mathcal{O}(T^{\nicefrac{3}{4}})$& \textsc{Conv-OPT} & \textbf{CF} \\
\citet{lekeufack2024optimistic} &$\mathcal{O}(\sqrt{(1+\mathcal{P}_T)T})$&\textbf{WD}& $\tilde{\mathcal{O}}({\sqrt{(1+\mathcal{P}_T)T}})$ & \textsc{Grad Eval} & \textbf{CF}, Known $\mathcal{P}_T$ \\ 
 \citet{cao2018online}  
      & $\mathcal{O}(\sqrt{(1+\mathcal{P}^\star_T)T})$&\textbf{WD}
      & $\mathcal{O}(T^{\nicefrac{3}{4}}(1+\mathcal{P}^\star_T)^{\nicefrac{1}{4}})$ 
       
      & \textsc{Conv-OPT} 
      & Known $\mathcal{P}_T^\star$ \\
\citet{liu2021simultaneously}  &$\mathcal{O}(\sqrt{(1+\mathcal{P}^\star_T)T})$&\textbf{WD}& $\mathcal{O}(\max(\sqrt{T},V_g))$ & \textsc{Conv-OPT} & Known $\mathcal{P}_T^\star$ \\
\textbf{This paper (Algorithm \ref{COCOAlgonewest1})}   & $\mathcal{O}(\sqrt{(1+\mathcal{P}_T(u_{1:T}))T})$& \textbf{UD} & $\mathcal{O}(\sqrt{(1 + \mathcal{P}_T^\star)T})$ & \textsc{PROJ} & --- \\
\textbf{This paper (Algorithm \ref{COCOAlgonew1})} & $\mathcal{O}((1+\mathcal{P}_T(u_{1:T}))\sqrt{T})$&\textbf{UD} & $\mathcal{O}(\max(T^{\nicefrac{3}{4}},\sqrt{T(1+\mathds{P}_T^\star)})$)  & \textsc{Grad Eval} & --- \\
\bottomrule
\end{tabular}
%\end{sc}
%\end{small}
%\end{center}
%\vskip -0.1in
}
\vspace{2 mm}
\caption{\small{Summary of the results on the COCO problem with dynamic regret and CCV bounds. Abbreviations have the following meanings- \textsc{Proj}: Euclidean projection on the time-varying constraint set, \textsc{Conv-OPT}: Solving a convex optimization problem over the decision set, \textsc{Grad Eval:} Evaluating the gradients of the cost and constraint functions, \textbf{UD}: Universal Dynamic Regret, $u_{1:T}:$ comparator action sequence for universal dynamic regret, \textbf{WD}: Worst-case Dynamic Regret, \textbf{CF}: Assumption \ref{com-feas} (Common feasibility), $\mathcal{P}_T:$ Known path length, $\mathcal{P}_T(u_{1:T}):$ path length of any arbitrary feasible comparator sequence}, $\mathcal{P}_T^\star:$ path length corresponding to the feasible minimizers of the constraint functions, $\mathds{P}_T^\star:$ minimum feasible path length.}
\label{comp-table}
\end{table*}

\paragraph{Convex Body Chasing:}

The COCO problem is closely related to the classic Convex Body Chasing (CBC) problem \citep{bubeck2020chasing, friedman1993convex, cbc}. In CBC, at each round $t \geq 1,$ the adversary first reveals a convex set $\mathcal{X}_t \subseteq \mathbb{R}^d$ and the learner is required to choose a feasible point $x_t \in \mathcal{X}_t$. The cost is measured by the total movement cost: $\sum\limits_{t=2}^{T}||x_t - x_{t-1}||.$ The performance of an online policy is measured in terms of the \emph{competitive ratio} metric which is  obtained by taking the ratio of the cost of the online algorithm and the cost of the optimal offline policy. \citet{cbc2} presents an online algorithm that achieves $O(\min(d, \sqrt{d \log T}))$ competitive ratio for $d$-dimensional decision space. COCO can be considered as a variant of CBC involving both cost and constraint functions, where the algorithm moves first in each round. Furthermore, in COCO, we are interested in additive performance guarantees (Regret, CCV), as opposed to multiplicative measures such as competitive ratio. See \citet[Section A.8]{sinha2024optimal} for further discussion on the connection between COCO and CBC. 

%One can see the similarity in the problem of COCO and the Convex Body Chasing(CBC) Problem. In COCO at every time a constraint function is revealed after the learner has played and the learner is expected to have been minimizing the losses subject to the constraint satisfaction. In CBC, it has been shown \cite{argue2019nearly, bubeck2020chasing, sellke2023chasing, przeslawski1989continuity} that choosing the Steiner point/ centroid is the optimal route to take. However the presence of the loss functions can skew the optimal point in COCO away from the Steiner points and towards the boundary. In that sense it is a generalization of the CBC problem.

\section{Problem Statement and Performance Metrics} \label{statement}
%\cmt{do it in the context of COCO}
We consider a sequential game played between an online learner and an adversary for $T$ rounds.  Let $\mathcal{X} \subseteq \mathbb{R}^d$ be a closed and convex decision set. For simplicity, we assume that the diameter of the decision set is finite\footnote{This assumption can be relaxed. See Remark \ref{param-free-remark}.} and upper-bounded by $D$. On the $t$\textsuperscript{th} round, the online learner plays an action $x_t$ from the decision set $\mathcal{X}.$ Upon observing the learner's action, the adversary reveals a convex \emph{cost function} $f_t : \mathcal{X} \mapsto \mathbb{R}$ and a closed and convex \emph{constraint function} $g_t : \mathcal{X} \mapsto \mathbb{R}.$ The constraint function encodes a constraint of the form $g_t(x) \leq 0.$
As a result, the learner incurs a cost of $f_t(x_t)$ and a constraint violation of $\max(0, g_t(x_t)) \equiv g_t^+(x_t)$ on round $t.$ The objective of the learner is to choose a sequence of actions so as to \emph{simultaneously} minimize the cumulative cost $\sum_{t=1}^T f_t(x_t)$ and the cumulative constraint violation (CCV) $\sum_{t=1}^T g_t^+(x_t)$. All cost and constraint functions are assumed to be $G$-Lipschitz with respect to a norm $||\cdot||$, which will be taken to be the standard Euclidean norm unless specified otherwise. The cost and constraint functions could be otherwise arbitrary and we do not impose any further restrictions on them.  
In particular, we do not make the common feasibility assumption (Assumption \ref{com-feas}), which has been a common limiting factor in previous works. %\cmt{cite and expand on it.}
\begin{remark}
In the special case when all constraint functions are identically equal to zero, we recover the standard Online Convex Optimization (OCO) framework \citep{hazan2022introduction}. 
\end{remark}

In the following, we introduce several definitions for the COCO problem that will be used extensively throughout the paper. These straightforwardly generalize the corresponding definitions in OCO. 

\begin{definition}[Comparator Sequence]
	A comparator (\emph{a.k.a.} benchmark) sequence denoted by $u_{1:T}\equiv \{u_1, u_2, \ldots, u_T\}$ is a sequence of actions in the decision set $\mathcal{X}$ with which the performance of an online algorithm is compared.
\end{definition}

As customary in the literature, the performance bounds of the proposed online algorithms will be given in terms of the \emph{path length} of the comparator sequence, defined below.

\begin{definition}[Path Length] \label{path-length-def}
    The path length of a comparator sequence $u_{1:T}$  is defined to be
       \begin{equation}
        \mathcal{P}_T(u_{1:T}) = \sum\limits_{t=2}^{T} ||u_t - u_{t-1}||.
    \end{equation}
\end{definition}

Intuitively, path length captures the degree of temporal variation of a comparator sequence. 

\begin{definition}[Feasible Set]
  The feasible set $\mathcal{X}^\star_t$ for round $t$ is defined to be the subset of the decision set $\mathcal{X}$ for which the $t$\textsuperscript{th} constraint is satisfied, \emph{i.e.,} \[\mathcal{X}^\star_t := \{x \in \mathcal{X} : \,g_t(x) \leq 0\}.\]
\end{definition}
Since the constraint functions are assumed to be closed and convex, the feasible sets are also closed and convex. Without any loss of generality, we assume the feasible sets to be non-empty.

\begin{definition}[Feasible Comparators]
	A comparator sequence $u_{1:T}$ is said to be feasible if each of its element satisfies the corresponding constraint, \emph{i.e.,} 
	\[u_t \in \mathcal{X}_t^\star, \forall t \geq 1.\]
\end{definition}

%Having defined these key concepts in the context of the standard OCO problem, we now turn towards the Constrained Online Convex Optimization problem. 

%The regret metrics of the COCO problem is defined in a slightly different way thann the standard OCO problem. In COCO we compare the performance of the algorithm against the comparators which lie in the constraint set. We explicitly define them to avoid confusion. 

%\begin{definition}[Static Regret of COCO]
%In the COCO problem, for any sequence of cost functions $f_1,\dots,f_T$ and any comparator $u$ in $\bigcap\limits_{t=1}^T \mathcal{X}^*_t$, the static regret is defined as
%\begin{equation}
%    \textrm{S-Regret}(u) = \sum\limits_{t=1}^{T}f_t(x_t) - \sum\limits_{t=1}^{T}f_t(u) 
%\end{equation}
%\end{definition}

\begin{definition}[Cumulative Constraint Violation (CCV)]
The cumulative constraint violation (CCV) at the end of round $T$ is defined as:
\begin{equation}\label{ccv-def}
	Q(T) = \sum_{t=1}^T (g_t(x_t))^+.
\end{equation} 
\end{definition}

\begin{definition}[Universal Dynamic Regret]
For any sequence of cost functions $f_{1:T}$ and any feasible comparator sequence $u_{1:T}$, the universal dynamic regret for an action sequence $x_{1:T}$ is defined as:
\begin{equation}\label{ud-reg-def}
    \textsc{UD-Regret}\big(f_{1:T};u_{1:T}\big) = \sum\limits_{t=1}^{T} f_t(x_t) - \sum\limits_{t=1}^{T} f_t(u_t) 
\end{equation}
\end{definition}
Note that the action sequence $x_{1:T}$ need not be feasible and can have a strictly positive CCV \eqref{ccv-def}. 
%We may drop the first argument from the above notation if the functions are clear from the context. 
Since a universal dynamic regret bound must hold for all feasible comparator sequences, which could possibly have widely varying path lengths, the algorithm must be oblivious of the comparator sequence and, specifically, of the path length of the comparator. In other words, the algorithm cannot tune its internal parameters according to the path lengths of any specific comparator sequence.

%We emphasize that the comparator path length $\mathcal{P}_T(u_{1:T})$ is unknown to the learner and may vary arbitrarily across feasible comparator sequences. Universal dynamic regret guarantees must therefore hold simultaneously for all feasible comparator sequences, without tuning algorithm parameters to any specific path length.
%\citet{zinkevich2003online} showed that the standard Online Gradient Descent (OGD) algorithm achieves a universal dynamic regret bound of $\mathcal{O}((1 + \mathcal{P}_T(u_{1:T}))\sqrt{T})$. Later this bound was improved to $\mathcal{O}(\sqrt{(1 + \mathcal{P}_T(u_{1:T}))T})$ by \cite{zhang2018adaptive} and shown to be minimax optimal. 

\begin{remark} %The notion of static regret can be shown to be a special case of the dynamic regret as defined above. Assume that the intersection of all feasible sets to be non-empty, \emph{i.e., } $\cap_{t=1}^T \mathcal{X}_t^\star \neq \emptyset.$ Let $x^\star$ belong to this non-empty intersection. 
Under the Common Feasibility assumption (Assumption \ref{com-feas}), there exists a common feasible action $x^\star \in \cap_{t=1}^T \mathcal{X}_t^\star.$ 
Hence, choosing this comparator every round, \emph{i.e.,} $u_t = x^\star, \forall t,$ we recover the static regret metric considered in earlier works \citep{sinha2024optimal, guo2022online, mahdavi2012trading}. 
 %In this settin g, \cite{sinha2024optimal} provides an optimal algorithm which simultaneously guarantees $\mathcal{O}(\sqrt{T)}$ static regret and $\tilde{\mathcal{O}}(\sqrt{T)}$ CCV. 
% As stated earlier, in this paper we focus on the universal dynamic regret metric.
\end{remark}
%
%\begin{remark}
%	Since a universal dynamic regret bound must hold for all feasible comparator sequences $u_{1:T}$, which could possibly have widely varying path lengths, the algorithm must be oblivious of the comparator sequence and, specifically, of the path length of the comparator. In other words, the algorithm cannot tune its internal parameters according to the path lengths of any specific comparator sequence.
%	
%	\end{remark}

\begin{definition}[Worst-case Dynamic Regret] \label{wdreg-def}
    For any sequence of cost functions $f_1,\dots,f_T$, the worst-case dynamic regret is defined as 
\begin{equation}
    \textsc{WD-Regret}(f_{1:T}) = \sum\limits_{t=1}^{T} f_t(x_t) -  \sum\limits_{t=1}^{T} \min_{x^\star_t \in \mathcal{X}^\star_t} f_t(x^\star_t).
\end{equation}
In other words, the worst-case dynamic regret metric considers a restricted class of benchmark sequences, each of which consists of instantaneous minimizers of the cost functions over the corresponding feasible sets. The minimum path length among all such benchmark sequences is denoted by $\mathcal{P}_T^\star.$ 
\end{definition}

\begin{remark} Unlike the universal dynamic regret, for the worst-case dynamic regret metric, we can assume, without much loss of generality, that a tight upper-bound to the path-length $\mathcal{P}_T^\star$ is known. This is because, in this case, the path-length can always be estimated within a constant factor via the standard doubling trick \citep{guo2022online, cao2018online}.  
\end{remark}
% In this paper, we look at the worst-case dynamic regret and the universal dynamic regret as performance benchmarks and design algorithms which provide guarantees for sublinear dynamic regret and CCV.

\paragraph{Comparison between Universal and Worst-Case Dynamic Regret Metrics:}
We now briefly explain the operational advantage of having a universal dynamic regret bound over a worst-case dynamic regret bound. Recall that our final objective is to select actions so as to minimize the cumulative cost $\sum_{t=1}^T f_t(x_t).$ 
A typical no-regret online algorithm, while competing against a feasible benchmark $u_{1:T}$, achieves a cumulative cost of 
\begin{eqnarray} \label{uni-dyn-reg}
	\sum_{t=1}^T f_t(x_t) \leq \underbrace{\sum_{t=1}^T f_t(u_t)}_{\textrm{term-I}} + \underbrace{\psi_T(\mathcal{P}_T(u_{1:T}))}_{\textrm{term-II}},
\end{eqnarray}  
where $\psi_T(\cdot)$ is some algorithm-dependent non-decreasing function of the path length of the comparators. To obtain the tightest possible guarantee on the cumulative cost, one must minimize the RHS of \eqref{uni-dyn-reg}. In case of worst-case dynamic regret, the comparator sequence is fixed to be $x_t^\star = \arg\min_{x \in \mathcal{X}_t^\star} f_t(x), t \geq 1,$ which minimizes only term-I in the upper bound \eqref{uni-dyn-reg}. However, the corresponding path length $\mathcal{P}_T^\star$ of this sequence may be large, resulting in a large value of term-II, and hence, a suboptimal bound on the cumulative cost. In contrast, under a universal dynamic regret guarantee, the bound \eqref{uni-dyn-reg} holds uniformly over \emph{all} feasible comparator sequences. This allows one to select a feasible benchmark sequence $u^\star_{1:T}$ that optimally balances the tradeoff between term-I and term-II, thereby yielding a strictly tighter cumulative cost bound. Similar observations have also been made in \citet{zhang2018adaptive}. 

\section{COCO Algorithm with Full Constraint Feedback}
%\cmt{Recover NeurIPS 24 result for static regret?}
%\cmt{Are we using the fact that the distance needs to be $2$-norm}?
%\cmt{Consider adding an illustrative figure of the original, auxiliary, and surrogate costs.}
%At every round, this algorithm requires the gradient information of the cost function and computing a projection on to the sub-level set of the constraint functions. 
In this Section, we present our first COCO algorithm, Algorithm \ref{COCOAlgonewest1}, that simultaneously guarantees a minimax optimal universal dynamic regret bound and a competitive CCV. Algorithm \ref{COCOAlgonewest1} is based on transforming the original cost functions so that infeasible actions are penalized in a geometrically meaningful way. The key idea is to add a distance-based penalty to the cost functions that ensures all global minimizers of the transformed cost function lie within the current feasible set, while preserving convexity and Lipschitz continuity. 

For any closed and convex set $S \subseteq \mathcal{X}$, let $\dist(x, S) \equiv \min_{y \in S}||x-y||$ denote the minimum distance from a point $x \in \mathcal{X}$ to the set $S,$ where the distance is measured with respect to the norm $||\cdot||.$ We now define the following auxiliary cost functions $\tilde{f}_t : \mathcal{X} \mapsto \mathbb{R}, t \geq 1$:  
\begin{eqnarray} \label{aux_fn_def}
	\ftilde_t(x) := f_t(x) + 2G\dist(x,\Xstart), ~~x \in \mathcal{X},
\end{eqnarray}
where, we recall, $G$ is an upper bound to the Lipschitz constant of the cost and constraint functions. 
Intuitively, the auxiliary cost function adds a penalty to the actions that are away from the current feasible set $\mathcal{X}_t^\star$. See Figure \ref{f-tilde-fig} in the Appendix for illustrative examples. The auxiliary cost functions will be used to construct the surrogate cost functions for which we would run a no-regret algorithm, as outlined in Algorithm \ref{COCOAlgogen}. 

 Since the function $\dist(\cdot, S)$ is known to be convex and $1$-Lipschitz \citep[Example 3.16]{boyd}, it follows that the auxiliary cost function $\tilde{f}_t$ is convex and $3G$-Lipschitz for any $t \geq 1.$ The following two lemmas underscore two important properties of the auxiliary cost functions, which will be useful later in the analysis. 

\begin{lemma}\label{lemma:9.1}
    For any feasible comparator sequence $u_{1:T}$ and any decision sequence $x_{1:T}$, we have $\textsc{UD-Regret}\big(f_{1:T};u_{1:T}\big)  \leq \textsc{UD-Regret}\big(\tilde{f}_{1:T};u_{1:T}\big),$ \emph{i.e.,}
    %the dynamic regret for the original cost functions $f_{1:T}$ is upper bounded by the dynamic regret for the auxiliary cost functions $\tilde{f}_{1:T},$ \emph{i.e.,}
    \[
        \sumlim_{t =1}^{T} \bigl(f_t(x_t) - f_t(u_t)\bigr) \leq \sumlim_{t=1}^T\bigl( \ftilde_t (x_t) - \ftilde_t(u_t)\bigr).
    \]
\end{lemma}
%\cmt{Consider writing the above statement in words in terms of Dynamic regret first, e.g., \[\textsc{D-Regret}(\{f_\tau(x_\tau)\}_{\tau=1}^t), \{u_\tau\}_{\tau=1}^t)) \leq \textsc{D-Regret}(\{\tilde{f}_\tau(x_\tau)\}_{\tau=1}^t), \{u_\tau\}_{\tau=1}^t)\] and then give the above inequality.}

\begin{proof}
  
  Definition \eqref{aux_fn_def} implies that for any $x_t \in \mathcal{X}$, we have  $\ftilde_t(x_t) \geq f_t(x_t).$ Furthermore, since $u_{1:T}$ is a feasible sequence, we have $u_t \in \mathcal{X}_t^\star, \forall t\geq 1.$ This yields $\dist(u_t, \mathcal{X}_t^\star) =0$, and hence, $\ftilde_t(u_t) = f_t(u_t), \forall t\geq 1.$ Combining the above two relations, the result follows.
  %and for any $x \in \Xstart, \ftilde_t(x) = f_t(x)$
\end{proof}
%\cmt{For easy readability, write in words what the following lemma means and then give the mathematical result.}
The next result states that for any round $t\geq 1,$ the minima of the auxiliary function $\tilde{f}_t$ over the entire decision set $\mathcal{X}$ lies in the feasible set $\mathcal{X}_t^\star$. See Section \ref{f-tilde-fig} in the Appendix for a visual illustration of this property. 
\begin{lemma}\label{lemma:9.2}
For any round $t \geq 1$, we have
    \[
    \arg\min_{x \in \mathcal{X}} \ftilde_t(x) \in  \mathcal{X}_t^\star .
    \]
\end{lemma}
\begin{proof}
%The proof proceeds via a contradiction argument. 
We prove the claim by means of contradiction by showing that any minimizer of $\tilde{f}_t$
 lying outside $\mathcal{X}_t^\star$ can be strictly improved by projecting it onto $\mathcal{X}_t^\star.$
 
    Let $u^\star \in \arg\min_{x \in \mathcal{X}} \ftilde_t(x)$, and for the sake of contradiction, assume that  $u^\star \notin \mathcal{X}_t^\star$. Let $v^\star := \mathsf{Proj}_{\mathcal{X}_t^\star} (u^\star)$ be the projection of $u^\star$ on to the set $\mathcal{X}_t^\star,$ \emph{i.e.,}
    \[ v^\star = \arg \min_{z \in \mathcal{X}_t^\star} || z- u^\star||.\]
     Since the set $\mathcal{X}_t^\star$ is closed and convex, the existence and uniqueness of the projection $v^\star$ are guaranteed by the Projection Theorem \citep[Proposition 2.2.1]{bertsekas2003convex}.  Thus we have the following series of inequalities:
    \begin{eqnarray*}
        \ftilde_t(u^\star) &\stackrel{(a)}{=}& f_t(u^\star) + 2G||v^\star-u^\star||\\
        &\stackrel{(b)}{\geq}& f_t(u^\star) + G||v^\star-u^\star|| + |f_t(v^\star) - f_t(u^\star)| \\
        &\geq& f_t(u^\star) + G||v^\star-u^\star|| + f_t(v^\star) - f_t(u^\star) \\
        &=& f_t(v^\star) + G||v^\star-u^\star|| \\
        &\stackrel{(c)}{>}& f_t(v^\star) \\
        &\stackrel{(d)}{=}& \ftilde_t(v^\star),
    \end{eqnarray*}
    where (a) follows from Definition \eqref{aux_fn_def}, (b) follows from the $G$-Lipschitzness of the cost functions, which yields $|f_t(v^\star) - f_t(u^\star)| \leq G||v^\star - u^\star||,$ (c) follows from our assumption that $u^\star \notin \mathcal{X}_t^\star$ and hence, $||v^\star - u^\star|| >0,$ and finally, $(d)$ follows from Definition \eqref{aux_fn_def} using the fact that $v^\star \in \mathcal{X}_t^\star.$ Hence, we have $\ftilde_t(u^\star) > \ftilde_t(v^\star)$. Since $u^\star$ is defined to be a minimizer of the auxiliary function $\ftilde_t$ over the entire decision set $\mathcal{X},$ the above constitutes a contradiction and concludes the proof. 
\end{proof}
%\cmt{Number each of the above steps and clearly explain them at the end.}

We now construct the $t$\textsuperscript{th} surrogate cost function $\hat{f}_t: \mathcal{X} \mapsto \mathbb{R}$ by adding the auxiliary cost function to the non-negative part of the constraint function, \emph{i.e.,}
\begin{eqnarray} \label{surr-cost-def}
\fhat_t(x) = g_t^+(x) + \ftilde_t(x), ~~ t \geq 1. 	
\end{eqnarray}
Clearly, the surrogate cost function $\hat{f}_t$ is convex and $4G$-Lipschitz as its Lipschitz constant $\mathsf{Lip}(\hat{f}_t)$ can be upper bounded as: \[\mathsf{Lip}(\hat{f}_t) \leq \mathsf{Lip}(g_t^+) + \mathsf{Lip}(\tilde{f}_t) \leq G+ \mathsf{Lip}(f_t) + 2G \mathsf{Lip}(\mathsf{dist}(\cdot)) \leq 4G.\] As stated earlier, our proposed algorithms follow the generic algorithmic template given by Algorithm \ref{COCOAlgogen}. For Algorithm \ref{COCOAlgonewest1}, we use the existing $\mathsf{ADER}$ algorithm, proposed by \citet{zhang2018adaptive}, as our OCO subroutine ($\mathcal{A}^{\textrm{OCO}}$) for minimizing the surrogate cost. %Recall that \mathsf{ADER} enjoys the following universal dynamic regret guarantee.
In our analysis, we will need only the final universal dynamic regret bound for $\mathsf{ADER}$ as given below.

\begin{theorem}[\citep{zhang2018adaptive}] \label{ader-reg-bd}
For any sequence of $G$-Lipschitz convex cost functions $\{\hat{f}_t\}_{t=1}^T$ defined on a convex domain with diameter $D$, the $\mathsf{ADER}$ algorithm achieves the following universal dynamic regret bound for any benchmark sequence $u_{1:T}:$ 
\[\textsc{UD-Regret}(\hat{f}_{1:T}; u_{1:T}) \leq \gamma \sqrt{T(1+\mathcal{P}_T(u_{1:T}))},\]
where $\gamma = O(GD\log \log T)$. 
\end{theorem}
In the following section, we will present a more general dynamic regret minimizing algorithm for problems with unbounded Lipschitz constants. Hence, we refer the reader to \citet{zhang2018adaptive} for details on the $\mathsf{ADER}$ algorithm. 
\begin{algorithm}
\caption{The First COCO Algorithm} \label{COCOAlgonewest1}
\begin{algorithmic}[1]
%    \STATE \textbf{Parameters}: $V=1,\Phi(x) = \exp(\lambda x) - 1,c=2G(D+1),\lambda = \frac{1}{2c \sqrt{1 + \mathcal{P}_T^\star} \sqrt{T}}$   
 %   \STATE \textbf{Initialization}: Set $x_1 = \mathbf{0}~, Q(0)=0$
% \REQUIRE  A generic OCO subroutine $\mathcal{A}^{\textrm{OCO}}$ with universal dynamic regret guarantees
% \STATE From the cost and constraint functions observed so far, construct a sequence of surrogate cost functions $\{\hat{f}_t\}_{t\geq 1}.$  
% \STATE Run $\mathcal{A}^{\textrm{OCO}}$ on the surrogate cost functions $\{\hat{f}_t\}_{t\geq 1}.$
\STATE Run Algorithm \ref{COCOAlgogen} with $\mathcal{A}^{\textrm{OCO}}=\mathsf{ADER} $ on the surrogate cost functions defined below: 
\begin{eqnarray} \label{surr-cost-def-alg2}
	 \hat{f}_t(x) = f_t(x) + g_t^+(x) + 2G \dist(x, \mathcal{X}_t^\star), ~x \in \mathcal{X}; ~~t \geq 1.
\end{eqnarray}

\end{algorithmic}
\end{algorithm}
We now state the main result of this section.

%\cmt{Write the result in the form of a theorem.}
\begin{theorem}[Universal Dynamic Regret and CCV for Algorithm \ref{COCOAlgonewest1}] \label{first-alg-bd}
	%Algorithm \ref{COCOAlgogen}, when run with any OCO algorithm with universal dynamic regret guarantee of the form $\tilde{O}(\sqrt{(1+\mathcal{P}_T(u_{1:T}))T})$ 
	Algorithm \ref{COCOAlgonewest1} achieves the following universal dynamic regret and CCV bounds:
	\begin{eqnarray*}
		\textsc{UD-Regret}\big(f_{1:T};u_{1:T}\big) \leq \zeta GD\sqrt{(1+\mathcal{P}_T(u_{1:T}))T}, ~~ \textsc{CCV}_T \leq  \zeta GD\sqrt{(1+\mathcal{P}_T^\star)T},
	\end{eqnarray*}
	where $u_{1:T}$ is any feasible comparator sequence, $\mathcal{P}_T^\star$ is the path length of the worst-case feasible comparator sequence given by $x_t^\star = \arg\min_{x \in \mathcal{X}_t^\star} f_t(x), ~t \geq 1,$ and $\zeta = O(\log \log T).$
\end{theorem}
%\subsection{Analysis}
\begin{proof}
The proof follows via a regret decomposition argument that simultaneously upper bounds the cumulative constraint violation and the universal dynamic regret using the regret guarantee of the OCO subroutine.

For any feasible comparator sequence $u_{1:T},$ we have the following dynamic regret decomposition equality:
\begin{eqnarray} \label{dyn-reg-decomp}
Q(T) + \textsc{UD-Regret}_T\big(\tilde{f}_{1:T};u_{1:T}\big)  &=&   Q(T) + \sumlim_{t=1}^T \bigl(\ftilde_t (x_t) - \ftilde_t(u_t)\bigr) \nonumber\\
       &=& \sum_{t=1}^T g^+_t(x_t)+\sumlim_{t=1}^T \bigl(\ftilde_t (x_t) - \ftilde_t(u_t)\bigr)\nonumber \\
        &\stackrel{(a)}{=}&\sumlim_{t=1}^T\bigl(\fhat_t(x_t) - \fhat_t(u_t)\bigr) \nonumber\\
        &=& \textsc{UD-Regret}_T\big(\hat{f}_{1:T};u_{1:T}\big),
   % &\leq& \gamma \sqrt{(1 + \patht{u_{1:t}})t}
\end{eqnarray}
where in step (a), we have used Definition \eqref{surr-cost-def} for the surrogate costs along with the feasibility of the comparator sequence. Finally, using Theorem \ref{ader-reg-bd}, the RHS of inequality \eqref{dyn-reg-decomp} can be upper bounded as: %\cmt{cite appropriate references}:
\begin{eqnarray} \label{surr-reg-bd45}
	 \textsc{UD-Regret}_T\big(\hat{f}_{1:T};u_{1:T}\big) \leq \gamma \sqrt{(1 + \mathcal{P}_T(u_{1:T})T},
\end{eqnarray}
where $\mathcal{P}_T(u_{1:T})$ is the path length of the benchmark sequence $u_{1:T}$, and $\gamma = O(GD\log \log  T).$

\paragraph{Bounding CCV:} Consider the comparator sequence $u_t:=x^\star_t, ~\forall t \in [T]$, where $x_t^\star$ is the global minimizer of the auxiliary cost function $\tilde{f}_t,$ \emph{i.e.,}  $x^\star_t = \arg\min_{x \in \mathcal{X}} \ftilde_t(x).$ By Lemma \ref{lemma:9.2}, the comparator sequence $x^\star_{1:T}$ is feasible. Furthermore, for any $t\geq 1,$ we have
\begin{eqnarray} \label{global-local}
	u_t=x_t^\star = \arg\min_{x \in \mathcal{X}}\tilde{f}_t(x) \stackrel{(a)}{=} \arg\min_{x\in \mathcal{X}_t^\star } \tilde{f}_t(x)\stackrel{(b)}{=} \arg\min_{x \in \mathcal{X}_t^\star} f_t(x), 
	\end{eqnarray}
where $(a)$ follows from Lemma \ref{lemma:9.2} and $(b)$ follows from the fact that $f_t(x)= \tilde{f}_t(x), \forall x \in \mathcal{X}_t^\star.$ Eqn.\ \eqref{global-local} shows that the comparator sequence $x^\star_{1:T}$ coincides with the worst-case benchmark used in Definition \ref{wdreg-def}. Let $\mathcal{P}_T^\star$ denote the minimum path length corresponding to the worst-case benchmarks. 
 
%The path length of this sequence $\mathcal{P}^*_t$ is a very commonly used quantity in the Online Learning literature to bound performance metrics of an algorithm in a dynamic setting. 

Since, by definition, $x_t^\star$ is a global minimizer of $\tilde{f}_t$, we have $\bigl( \ftilde_t (x_t) - \ftilde_t(x^\star_t)\bigr) \geq 0, \forall t.$ This implies that $\textsc{UD-Regret}\big(\tilde{f}_{1:T};x^\star_{1:T}\big) \geq 0.$ Hence, from the dynamic regret decomposition inequality \eqref{dyn-reg-decomp}, we obtain the following upper bound to CCV:
\begin{equation*}
    Q(T) \leq \textsc{UD-Regret}_T\big(\hat{f}_{1:T};x^\star_{1:T}\big) \leq \gamma \sqrt{(1 + \mathcal{P}^\star_T)T}.
\end{equation*}

\paragraph{Bounding Regret:} Finally, let $u_{1:T}$ be any feasible comparator sequence. Hence, we have  
\begin{eqnarray*}
    \textsc{UD-Regret}_T\big(f_{1:T};u_{1:T}\big) &\stackrel{(a)}{\leq}& \textsc{UD-Regret}_T\big(\tilde{f}_{1:T};u_{1:T}\big) \\
     &\stackrel{(b)}{\leq}& \textsc{UD-Regret}_T\big(\hat{f}_{1:T};u_{1:T}\big) \\
     &\stackrel{(c)}{\leq}& \gamma \sqrt{(1 + \mathcal{P}_T({u_{1:T}}))T}.
\end{eqnarray*}
where (a) follows from Lemma \ref{lemma:9.1}, (b) follows from Eqn.\ \eqref{dyn-reg-decomp}, where we have used the non-negativity of $Q(T),$ and (c) follows from  Eqn.\ \eqref{surr-reg-bd45}. 
\end{proof}
%\cmt{Include a discussion on the proof technique (may be at the beginning) including how it differs from \cite{sinha2024optimal}, e.g., different choice of the surrogate costs that does not depend on $Q(t)$, no need for adaptive subroutines etc.}
\begin{remark}
	It is interesting to note that the above proof remains unaltered even when the $\dist(\cdot, \mathcal{X}_t^\star)$ term in the auxiliary cost function \eqref{aux_fn_def}
	 is defined with respect to \emph{any} arbitrary norm. This allows one to obtain tighter regret bounds with improved dependence on the ambient dimension by exploiting the geometry of the cost and constraint functions. 
\end{remark}
\begin{remark}
The $G$-Lipschitzness of the constraint functions imply that $g_t^+(x) \leq G \dist(\cdot, \mathcal{X}_t^\star), \forall x \in \mathcal{X}.$ Hence, intuitively, both $g_t^+(\cdot)$ and $\dist(\cdot, \mathcal{X}_t^\star)$ measure some form of ``distance'' to the current feasible set $\mathcal{X}_t^\star$. This raises a natural question: \emph{Can we drop the last term in Eqn.\ \eqref{surr-cost-def-alg2} involving the $\dist(\cdot, \cdot)$ function and simply use $f_t + g_t^+(x)$ as the surrogate cost function?} The answer to this question turns out to be negative. See Appendix \ref{dist-nec} for a counter-example.
\end{remark}
\begin{remark} \label{param-free-remark}
	Algorithm \ref{COCOAlgonewest1} can be made parameter-free w.r.t. the horizon-length $T$ by using the standard doubling trick. On the other hand, $\mathsf{ADER}$ uses the knowledge of the diameter of the decision set $(D)$. Hence, it is not a priori clear how to make the algorithm parameter-free w.r.t. $D.$ However, \citet[Appendix J, Algorithm 6 and Lemma 10]{jacobsen2022parameter} give a simple reduction scheme that transforms any online algorithm with a dynamic regret bound on a bounded domain into one achieving a similar bound on the unbounded domain. This reduction uses an additional parameter-free 1 dimensional OCO subroutine (which can be constructed, \emph{e.g.,} using the coin-betting scheme of \citet[Chapter 9]{orabona2019modern}). Using this scheme, the OCO subroutine $\mathsf{ADER}$, and hence, Algorithm \ref{COCOAlgonewest1} can be made parameter-free with respect to $D$ in a black-box fashion.
\end{remark}

\paragraph{Extension to quasiconvex constraint functions:}
 Algorithm~\ref{COCOAlgonewest1} can be extended to the setting where the constraint functions are assumed to be only \emph{quasiconvex} and 
$G$-Lipschitz, rather than convex and $G$-Lipschitz, as assumed earlier. Recall that a function $g: \mathcal{X} \to \mathbb{R}$ is called \emph{quasiconvex} if its domain is convex and all its sublevel sets, given by  
\[ S_\alpha = \{x \in \mathcal{X}: g(x) \leq \alpha \}, ~\alpha \in \mathbb{R},\]
are convex. Clearly, all convex functions are quasiconvex. See \citet[Section~3.4]{boyd} for standard examples of non-convex yet quasiconvex functions. 

Given a quasiconvex constraint function $g_t,$ let $\mathcal{X}_t^\star = \{ x \in \mathcal{X}: g_t(x) \leq 0\}$ denote its feasible set. Since $\mathcal{X}_t^\star$ is convex, we now define a new $G$-Lipschitz convex constraint function $h_t : \mathcal{X} \mapsto \mathbb{R}_+$ as follows: \[h_t(x) = G\dist(x, \mathcal{X}_t^\star), ~x \in \mathcal{X}.\]
Clearly, $g_t$ and $h_t$ have the same feasible set. We then run Algorithm~\ref{COCOAlgonewest1} using the original cost function $f_t$ and the new convex constraint function $h_t$ for all $t \geq 1.$ Using the Lipschitz continuity of 
$g_t,$ we have $g_t^+ \leq h_t(x), \forall x \in \mathcal{X}.$  Consequently, the regret and cumulative constraint violation bounds established in Theorem~\ref{first-alg-bd} continue to hold in this quasiconvex setting.

\paragraph{Computational aspects:} Note that the universal dynamic regret minimizer subroutine $\mathsf{ADER}$ requires only a sub-gradient of the surrogate cost function as an input on every round. From Eqn.\ \eqref{surr-cost-def-alg2}, it follows that in order to compute a sub-gradient of the surrogate cost function $\hat{f}_t$, we need to compute sub-gradients of the cost and constraint functions and the gradient of the function $\dist(\cdot, \mathcal{X}_t^\star).$ To compute the latter, let $\mathsf{Proj}_{\mathcal{X}_t}(x_t)$ denote the Euclidean projection of $x_t$ onto the feasible set $\mathcal{X}_t^\star.$ Then we have 
\begin{eqnarray*}
	\nabla_x \dist(x, \mathcal{X}_t^\star)|_{x=x_t} = \nabla_x \min_{y \in \mathcal{X}_t^\star} ||x-y|||_{x=x_t}= \frac{x_t-\mathsf{Proj}_{\mathcal{X}_t}(x_t)}{||x_t-\mathsf{Proj}_{\mathcal{X}_t}(x_t)||},
\end{eqnarray*}
which is precisely the unit vector in the direction from $x_t$ to its projection on to $\mathcal{X}_t^\star.$ Thus the main computational bottleneck for running Algorithm \ref{COCOAlgonewest1} is the computation of projections on the time-varying feasible sets $\mathcal{X}_t^\star$'s on each round. Note that the universal dynamic regret minimizer subroutine ($\mathsf{ADER}$) also involves a projection onto the (fixed) decision set $\mathcal{X}.$ However, in most practical cases, $\mathcal{X}$ has simple forms (\emph{e.g.,} Euclidean box, $d$-dimensional ball), which enable fast computation of this projection step. Hence, the run time of Algorithm \ref{COCOAlgonewest1} is primarily dominated by the former projection step.

%\cmt{Computing the gradient of the distance function.}
\section{COCO with First-order Feedback}
While Algorithm \ref{COCOAlgonewest1} achieves the optimal dynamic regret guarantee, projecting onto time-varying sets may be computationally prohibitive in high-dimensional settings. This motivates us to propose a new \emph{projection-free} algorithm, Algorithm \ref{COCOAlgonew1}, that relies only on the first-order information of the cost and constraint functions. Compared to the previous algorithm, Algorithm \ref{COCOAlgonew1} avoids the costly projection step onto the time-varying feasible sets altogether. Although it features a sub-optimal dependence on the path length in its regret bound, we will see that Algorithm \ref{COCOAlgonew1} achieves a tighter constraint violation guarantee in rapidly varying environments. Readers willing to assume the adaptive regret bound in Theorem \ref{meta_dyn_reg_ogd2} can skip Section \ref{prelims} and proceed directly to Section \ref{ada} without much loss of continuity.
%Furthermore, to compute the projections, Algorithm \ref{COCOAlgonewest1} requires to know the entire constraint function after each round, whereas Algorithm \ref{COCOAlgonew1} needs only the gradients of the cost and constraint functions. 

\subsection{Preliminaries: Universal Dynamic Regret with Unbounded Lipschitz Constants}  \label{prelims}
Algorithm \ref{COCOAlgonew1} uses the common template as given by Algorithm \ref{COCOAlgogen}. However, as we will see in the sequel, the Lipschitz constants of the surrogate cost functions in this case cannot be upper bounded \emph{a priori} as their growth depends on the past actions of the algorithm. This must be contrasted with the surrogate cost functions \eqref{surr-cost-def} used by Algorithm \ref{COCOAlgonewest1}, which are $4G$-Lipschitz irrespective of the history.
Hence, as a prerequisite, we first design an adaptive OCO subroutine, called $\mathsf{AHAG}$, that yields a universal dynamic regret bound for any sequence of convex cost functions with \emph{unbounded} Lipschitz constants. This is achieved by combining logarithmically many base policies, each running an instance of the $\mathsf{AdaGrad}$ sub-routine \citep{duchi2011adaptive} with different path-length estimates, with the $\mathsf{AdaHedge}$ algorithm \citep{de2014follow}. See Algorithm \ref{ogd_alg} for the pseudocode of $\mathsf{AdaGrad}$. Note that $\mathsf{ADER}$ achieves the minimax-optimal $\mathcal{O}(\sqrt{(1 + \mathcal{P}_T)T})$ universal dynamic regret bound. However, $\mathsf{ADER}$ requires the range and the gradient norms of all cost functions to be bounded by a constant \citep[Assumptions 1 and 2]{zhang2018adaptive}, and hence, it cannot be applied in our setting. To the best of our knowledge, the $\mathsf{AHAG}$ subroutine, described in Algorithm \ref{alg:AdaHedgeGrad}, is new and might be of independent interest. 
%In our analysis of COCO, we pass carefully constructed surrogate loss functions to a subroutine and we do not have any constant upper bound for these surrogate loss functions that is known a-priori. 

%For our second algorithm, we construct a sequence of surrogate loss functions $\{\hat{f}_t\}_{t\geq 1}$ which depend on the prior actions of the algorithm in a recursive fashion. As a result, we will see that the Lipschitz constant of $\hat{f}_t$ can not be upper bounded effectively at the beginning and we need to employ an adaptive OCO subroutine that $\mathcal{A}^{\textrm{OCO}}$ that adapts to the time-varying Lipschitz constants on-the-go. 

%We construct the above online policy in two steps. 
In Theorem \ref{dyn_reg_ogd}, we give two different adaptive dynamic regret bounds for the $\mathsf{AdaGrad}$ subroutine, corresponding to known and unknown path lengths. These results will be tightened later in Theorem \ref{meta_dyn_reg_ogd2} using multiple experts. While the dynamic regret of the $\mathsf{AdaGrad}$ algorithm for strongly-convex loss functions has been investigated in the literature \citep{nazari2024dynamic}, its dynamic regret bound for the basic convex case still remains under-explored.  
%For the worst-case dynamic regret guarantee, Algorithm \ref{ogd_alg} tunes the step sizes according to the path length $\mathcal{P}_T^\star$ corresponding to the worst-case benchmark. However, the universal dynamic regret bound does not assume any knowledge of the path length of the comparator. 
%\subsection{Preliminaries: Dynamic Regret Bounds for Adaptive OGD}
%We defer the proof to Appendix \ref{dyn_reg_ogd_proof}. 

% Although this bound can be obtained from the standard proof techniques \citep[Theorem 3]{zinkevich2003online}, we state and prove it here for easy reference.  

\begin{algorithm}
\caption{$\mathsf{AdaGrad}$: Online Gradient Descent with Adaptive step sizes} \label{ogd_alg}
\begin{algorithmic}[1]
    \STATE \textbf{Input} : Convex decision set $\mathcal{X}$, sequence of convex cost functions $\{\hat{f}_t\}_{t=1}^{T}, \textrm{diam}(\mathcal{X}) = D,$  $\mathsf{Proj}_{\mathcal{X}}(\cdot)=$ Euclidean projection on the convex set $\mathcal{X}$, non-increasing step sizes $\{\eta_t\}_{t \geq 1}$ .
    \STATE \textbf{Initialize} : $x_1 \in \mathcal{X}$ arbitrarily.
   % \State \textbf{Parameters}: Adaptive step size sequence \[\eta_t = \frac{\sqrt{2} D}{2 \sqrt{\sum_{\tau=1}^t ||\nabla_\tau||_2^2}}, t\geq 1\] 
    %\State \textbf{Intitialization}: Set $x_1 \in \mathcal{X}, Q(0)=0$
    \FOR{$t=1:T$}
        \STATE Play $x_t$ and compute $\nabla_t = \nabla \hat{f}_t(x_t)$
        \STATE Set $x_{t+1} = \mathsf{Proj}_{\mathcal{X}}(x_t - \eta_t \nabla_t)$  
    \ENDFOR
    \end{algorithmic}
\end{algorithm}

% Let $\{\hat{f}_t\}_{t \geq 1},$ be a sequence of convex cost functions and $\{x_t^\star\}_{t \geq 1}$ any feasible comparator sequence. Then the dynamic regret (against the comparator sequence $\{x_t^\star\}_{t=1}^T$) of any online policy which produces the feasible action sequence $\{x_t\}_{t \geq 1}$ is defined to be: 
% \begin{eqnarray} \label{dyn-reg-def}
% 	\textrm{D-Regret}_T \equiv \sum_{t=1}^T \hat{f}_t (x_t) - \sum_{t=1}^T \hat{f}_t(x_t^\star). 
% \end{eqnarray}
% We define the path length $\mathcal{P}_T$ of a comparator sequence $\{x_t^\star\}_{t=1}^T$) as the Euclidean length of the path connecting the comparator action sequence, \emph{i.e.,} 
% \begin{eqnarray} \label{path-len-def}
%  \mathcal{P}_T \equiv  \sum\limits_{t=1}^{T-1} ||x_{t+1}^\star - x_{t}^\star||.
% \end{eqnarray}
% Note that for the weaker static regret metric where the comparator actions are time-invariant, the path-length is zero \citep{hazan2022introduction}.

%The following theorem gives an upper bound to the dynamic regret of the adaptive OGD policy given above in Algorithm \ref{ogd_alg}.
% Although different flavour of dynamic regret bounds with different algorithms are known in the literature \citep{pmlr-v38-jadbabaie15}, we state and prove this particular result for the \mathsf{AdaGrad} policy given in Algorithm \ref{ogd_alg}.  
%give a simple proof of it in the Appendix for an easy reference. 
\begin{theorem}[Dynamic Regret Bounds for $\mathsf{AdaGrad}$] \label{dyn_reg_ogd}
	The $\mathsf{AdaGrad}$ subroutine, given in Algorithm \ref{ogd_alg}, achieves the following adaptive \textbf{dynamic regret} bound, denoted by $\textsc{D-Regret}(\hat{f}_{1:T};\mathcal{P}_T)$, for any sequence of convex cost functions $\hat{f}_{1:T}$ and any comparator sequence $u_{1:T}$ whose path length is \underline{known} to be at most $\mathcal{P}_T$, using an adaptive learning rate sequence $\eta_t = \frac{(D+1) \sqrt{1+\mathcal{P}_T}}{ \sqrt{2\sum_{\tau=1}^t ||\nabla_\tau||^2}}, ~t\geq 1:$ 
	\begin{eqnarray} \label{d-regret-bd}
		     \textsc{D-Regret}(\hat{f}_{1:T};\mathcal{P}_T) \leq 
   (D+1) \sqrt{2(1+\mathcal{P}_T)}
    \sqrt{\sum\limits_{t=1}^T ||\nabla_t||^2}.
	\end{eqnarray}
    $\mathsf{AdaGrad}$ also achieves the following \textbf{universal dynamic regret} bound, using the path-length-independent adaptive step size sequence $\eta_t = \frac{(D+1)}{ \sqrt{2\sum_{\tau=1}^t ||\nabla_\tau||^2}}, ~t \geq 1:$ 
	\begin{eqnarray} \label{ud-regret-bd}
		     \textsc{UD-Regret}(\hat{f}_{1:T}; u_{1:T}) \leq 
   \sqrt{2}(D+1) \big(1+\mathcal{P}_T(u_{1:T})\big)
    \sqrt{\sum\limits_{t=1}^T ||\nabla_t||^2},
	\end{eqnarray} 
    where $\nabla_t \equiv \nabla \hat{f}_t(x_t), \forall t \geq 1$ and $\textrm{diam}(\mathcal{X})=D.$
    % and $\mathcal{P}_T \equiv  \sum\limits_{t=1}^{T-1} ||x_{t+1}^\star - x_{t}^\star||$ is the given upper bound on the path-length of the comparator sequence. 
\end{theorem}
Notably, Theorem \ref{dyn_reg_ogd} holds without assuming any uniform upper bound for the Lipschitz constants of the cost functions.
Our proof of Theorem \ref{dyn_reg_ogd} extends the static regret analysis of the $\mathsf{AdaGrad}$ algorithm from \citet[Theorem 4.14]{orabona2019modern} to the dynamic regret setting.  See Appendix \ref{dyn_reg_ogd_proof} for the details.
% In the above, we assume that the path length (or a good upper bound to it) of the comparator sequence $\mathcal{P}_T$ is known. This is not a limiting assumption for the case of bounding the \emph{worst-case} dynamic regret. The path length of the worst-case comparators can be computed after the constraint function has been revealed. Allowing us to apply the standard doubling trick (\cite{cesa2006prediction}) to estimate the path length of the worst case comparators. Note however this is infeasible for the general case of universal dynamic regret. Also note that this requires the full knowledge of the constraint functions up to the current time. 

%Note that the dynamic regret bound above is not optimal. It can be improved to $O(\sqrt{P_T T})$ \cite{zhang2018adaptive}. However, this bound is insufficient for our purpose as it is not data-dependent and assumes a pre-specified upper-bounds on the gradient norms. 
%\iffalse
\subsubsection{Tighter Universal Dynamic Regret Bound with Multiple Experts} \label{tightened-adagrad}

Theorem \ref{dyn_reg_ogd} states that the standard $\mathsf{AdaGrad}$ subroutine achieves a suboptimal universal dynamic regret bound proportional to the path length of the comparator sequence. However, the same theorem also shows that if the path length of the comparator $u_{1:T}$ is known, $\mathsf{AdaGrad}$ (with a different step size sequence) achieves a tighter dynamic regret bound, which scales with the square root of the path length. In this section, we obtain an optimal universal dynamic regret bound by combining different $\mathsf{AdaGrad}$ sub-routines, each corresponding to different path length estimates, through the $\mathsf{AdaHedge}$ algorithm \cite[Section 7.6]{orabona2019modern}, improving the dependence on the path length to $\sqrt{1+\mathcal{P}_T(u_{1:T})}.$ 
The $\mathsf{AdaHedge}$ algorithm, introduced by \cite{de2014follow}, is a generalization of the celebrated $\mathsf{Hedge}$ algorithm for the classic Prediction with Expert advice problem \citep{cesa2006prediction}. $\mathsf{AdaHedge}$ is adaptive to the Lipschitz constants of the cost functions in the sense that, unlike $\mathsf{Hedge},$ it does not need to know a uniform upper bound to the Lipschitz constants of the costs \emph{a priori}. See Theorem \ref{adareg} in the Appendix for the regret bound for the $\mathsf{AdaHedge}$ algorithm. Using ideas similar to \citet{zhang2018adaptive}, our $\mathsf{AHAG}$ subroutine, described in Algorithm \ref{alg:AdaHedgeGrad}, runs an expert tracking $\mathsf{AdaHedge}$ algorithm, where each expert runs an instance of $\mathsf{AdaGrad}$ subroutine with step sizes tuned to their own guess of the path length. 
%This idea is similar to \textsc{ADER} algorithm \citep{zhang2018adaptive} which uses  
%\textsc{AdaHedge} is our meta-algorithm to track the experts using the \mathsf{AdaFTRL} algorithm first introduced in \cite{orabona2018scale}. 
%The new algorithm, called $\mathsf{AHAG}$, is described in Algorithm \ref{alg:AdaHedgeGrad}. 

%We give a data-dependent bound for the universal dynamic regret which is adaptive to the $\ell_\infty$ norm of the loss functions.

% \begin{algorithm}[H]
% \caption{\mathsf{AdaHedge}}
% \label{alg:adahedge}
% \begin{algorithmic}[1]
% \Require $\alpha > 0$
% \State $\lambda_1 = 0$
% \State $x_1 = [1/d, \ldots, 1/d] \in \mathbb{R}^d$
% \State $\theta_1 = 0 \in \mathbb{R}^d$
% \For{$t = 1$ to $T$}
%     \State Output $x_t$
%     \State Receive $g_t \in \mathbb{R}^d$ and pay $\langle g_t, x_t \rangle$
%     \State Update $\theta_{t+1} = \theta_t - g_t$
%     \State Set $\delta_t = 
%     \begin{cases}
%         \langle g_1, x_1 \rangle - \min_{j=1,\ldots,d} g_{1,j}, & t = 1 \\
%         \lambda_t \ln \left( \sum_{j=1}^d x_{t,j} \exp(-g_{t,j}/\lambda_t) \right) + \langle g_t, x_t \rangle, & \text{otherwise}
%     \end{cases}$
%     \State Update $\lambda_{t+1} = \lambda_t + \frac{\alpha}{\delta_t}$
%     \State Update $x_{t+1,j} \propto \exp\left( \frac{\theta_{t+1,j}}{\lambda_{t+1}} \right), \quad j = 1, \ldots, d$
% \EndFor
% \end{algorithmic}
% \end{algorithm}

\begin{algorithm}[h]
\caption{$\mathsf{AHAG}$: $\mathsf{AdaHedge}$ with $\mathsf{AdaGrad}$ Algorithm}
\label{alg:AdaHedgeGrad}
\begin{algorithmic}[1]
\STATE\textbf{Initialize:} $N = \lceil\frac{1}{2}\log_2( 1 + DT)\rceil+1$ experts. Expert $i \in [N]$ runs $\mathsf{AdaGrad}$ (Algorithm \ref{ogd_alg}) with the following adaptive step size sequence
\begin{equation}
\eta^i_t = \frac{(D+1) 2^{i-1}}{\sqrt{2\sum\limits_{\tau = 1}^t ||\nabla f_\tau(x_\tau^i)||^2}}, ~~t \geq 1.    
\end{equation}

Initialize the weight vector to the uniform distribution $w_1 \gets [1/N, \ldots, 1/N]$; Choose the initial actions of the experts $(x_1^i, \forall i)$ arbitrarily from $\mathcal{X}$
%\STATE $X_1= \mathbf{0} \in \mathbb{R}^{N\times N}$
%\STATE $x_1 =  \in \mathbb{R}^N$
\FOR{$t = 1:T$}
    \STATE Play a convex combination of experts actions with the coefficients given by the weight vector $w_t:$ \[x_t = \sum_{i=1}^N w_t^ix_t^i .\]
    \STATE Recieve the current cost function $f_t$ (only gradient information suffices).
    \FOR{experts $i = 1$ to $N$}
        \STATE Pass $f_t$ to their $\mathsf{AdaGrad}$ subroutine (Algorithm \ref{ogd_alg}) with learning rate $\eta^i_t$
        \STATE Receive the next action $x_{t+1}^i$
       % \STATE Set the $i^\text{th}$ column $[X_{t+1}]_i = x_{t+1}^i$
    \STATE Compute the loss of the $i$\textsuperscript{th} expert \[
    h^i_{t+1} = \big(\langle\nabla f_t(x^i_{t+1}),x^i_{t+1}\rangle, ~i \in [N]\big) \]
     \ENDFOR
    \STATE Pass loss vector $h_{t+1}=\big(h^i_{t+1}, i\in [N] \big)$ to $\mathsf{AdaHedge}$ and obtain the updated weight vector $w_{t+1}$.
\ENDFOR
\end{algorithmic}
\end{algorithm}

$\mathsf{AHAG}$ is a first-order algorithm, and its main computational bottleneck is the computation of $N$ gradients in each round (line $8$).
The following theorem gives a universal dynamic regret bound for $\mathsf{AHAG}$, which will be used as the no-regret subroutine $\mathcal{A}^{\textrm{OCO}}$ in Algorithm \ref{COCOAlgonew1}. 
\begin{theorem} \label{meta_dyn_reg_ogd2}
The $\mathsf{AHAG}$ algorithm, given in Algorithm \ref{alg:AdaHedgeGrad}, achieves the following Lipschitz-adaptive universal dynamic regret bound for any cost functions $f_{1:T}$ and any benchmark sequence $u_{1:T}:$  
	\begin{equation}\label{meta-regret-bd}
	\begin{aligned}	    
		   %  &\quad \textsc{UD-Regret}_T (u_{1:T}) \leq 
   %5(D+1)\sqrt{4 + \ln N} \sqrt{1+\mathcal{P}_T(u_{1:T})}
    %\sqrt{\sum\limits_{t=1}^T ||\nabla_t||^2},
    &\quad \textsc{UD-Regret}(f_{1:T}; u_{1:T}) \leq 
   \gamma \sqrt{1+\mathcal{P}_T(u_{1:T})}\sqrt{\sum\limits_{t=1}^T ||\nabla_t||^2},
	\end{aligned}
	\end{equation} 
    where $\nabla_t \equiv \nabla f_t(x_t), \forall t \geq 1,$ and $\gamma \equiv O(D \log \log T),$ where the decision set $\mathcal{X}$ is bounded within an Euclidean ball of radius at most $D/2$.
    %and $N = \frac{1}{2}\log_2(\lceil 1 + DT\rceil)+1.$
    % and $\mathcal{P}_T \equiv  \sum\limits_{t=1}^{T-1} ||x_{t+1}^\star - x_{t}^\star||$ is the given upper bound on the path-length of the comparator sequence. 
\end{theorem}
\subsection{The Second COCO Algorithm: Design and Analysis} \label{ada}
In this section, we present our second COCO algorithm that achieves competitive performance for any convex and bounded decision set while utilizing only first-order gradient feedback. 
%This must be contrasted with Algorithm \ref{COCOAlgonewest1}, which needs to know the full constraint functions on each round. 
% This is achieved by using the \mathsf{AHAG} policy given in Algorithm \ref{alg:AdaHedgeGrad} to learn the surrogate loss functions $\hat{f}_t^i$ as its input to bound the $\textrm{D-Regret}'_t(x^*_{1:t})$ term in \eqref{regretdecompositioninequality}. 
The pseudocode of the proposed algorithm is given below in Algorithm \ref{COCOAlgonew1}.
%We invoke Theorem \ref{meta_dyn_reg_ogd} to get the following.
\iffalse
\begin{algorithm}[h]
\caption{The Second COCO Algorithm} \label{COCOAlgo2}
\begin{algorithmic}[1]
    \STATE \textbf{Parameters}: $\Phi(x) = x^2~,V=50G^2(D+1)^2(4 + \ln N)~\sqrt{T}$
    \STATE \textbf{Initialization}: Set $x_1 = \mathbf{0}~, Q(0)=0$
    \FOR{$t=1:T$}
        \STATE Play $x_t,$ observe $f_t$ and $g_t$  
        \STATE $Q(t) \leftarrow Q(t-1) + \max(g_t(x_t),0)$  
        \STATE Compute $\hat{f}_t$ acccording to \eqref{surrogate}
        \STATE Pass $\hat{f}$ to $\mathsf{AHAG}$, described in Algorithm \ref{alg:AdaHedgeGrad}.
        \STATE Receive $x_{t+1}$ from the $\mathsf{AHAG}$ algorithm.
    \ENDFOR

\end{algorithmic}
\end{algorithm}
\fi

\begin{algorithm}
\caption{The Second COCO Algorithm (Projection-free)} \label{COCOAlgonew1}
\begin{algorithmic}[1]
%    \STATE \textbf{Parameters}: $V=1,\Phi(x) = \exp(\lambda x) - 1,c=2G(D+1),\lambda = \frac{1}{2c \sqrt{1 + \mathcal{P}_T^\star} \sqrt{T}}$   
 %   \STATE \textbf{Initialization}: Set $x_1 = \mathbf{0}~, Q(0)=0$
% \REQUIRE  A generic OCO subroutine $\mathcal{A}^{\textrm{OCO}}$ with universal dynamic regret guarantees
% \STATE From the cost and constraint functions observed so far, construct a sequence of surrogate cost functions $\{\hat{f}_t\}_{t\geq 1}.$  
% \STATE Run $\mathcal{A}^{\textrm{OCO}}$ on the surrogate cost functions $\{\hat{f}_t\}_{t\geq 1}.$
\STATE \textbf{Parameters:} %$V=50G^2(D+1)^2(4 + \ln N)~\sqrt{T},$ 
$V=\tilde{O}(GD\sqrt{T}), Q(0) =0,$
Quadratic Lyapunov function $\Phi(x)=x^2.$
\STATE Run Algorithm \ref{COCOAlgogen} with $\mathcal{A}^{\textrm{OCO}}=\mathsf{AHAG}$ (Algorithm \ref{alg:AdaHedgeGrad}), with the surrogate cost functions defined as follows:
\begin{eqnarray}\label{surr-cost23}
	 \hat{f}_t(x) = Vf_t(x) + \Phi'(Q(t))g_t^+(x), ~x \in \mathcal{X}; ~~t \geq 1,
\end{eqnarray}
%where we choose the parameter  
The variable $Q(t)$ denotes the cumulative CCV up to round $t,$ which evolves as follows:
\begin{eqnarray} \label{q-ev-new}
	Q(t)= Q(t-1)+g_t^+(x_t), ~~t \geq 1.
\end{eqnarray}
\end{algorithmic}
\end{algorithm}

\paragraph{Analysis:} 
Let $\Phi: \mathbb{R} \mapsto \mathbb{R}$ be a non-decreasing convex function ($\Phi(0)=0$), which will be used as a Lyapunov (\emph{a.k.a.} potential) function in our analysis. 
Since the function $\Phi(\cdot)$ is convex, for any two real numbers $x, y,$ we have $\Phi(x) - \Phi(y) \leq  \Phi'(x)(x-y)$. 
%We choose this as our lyapunov function to construct a surrogate loss function. First note that this follows from the previous line 
Let the variable $Q(t)$ denote the CCV up to round $t,$ which evolves as $Q(t)=Q(t-1)+g_t^+(x_t)$. Thus, we can upper bound the increase in the Lyapunov function on round $t$ as:
\begin{eqnarray}\label{phi-eq}
\Phi\bigl(Q(t)\bigr) - \Phi\bigl(Q(t-1)\bigr) \leq \Phi'(Q(t))g_t^+(x_t). 
\end{eqnarray}
Let $u_{1:T}$ be any feasible comparator sequence and $V$ be a positive parameter to be fixed later. Adding $V\bigl(f_t(x_t) - f_t(u_t)\bigr)$ to both sides of inequality \eqref{phi-eq} and summing over $t = 1$ to $T$, we obtain
\[
\Phi\bigl(Q(T)\bigr) + V\sum\limits_{t=1}^T \Bigr( f_t(x_t) - f_t(u_t)\Bigl) \leq \sum\limits_{t=1}^T \Phi'(Q(t))g_t^+(x_t) + V\sum\limits_{t=1}^T \Bigr( f_t(x_t) - f_t(u_t)\Bigl).
\]
Using the feasibility of the comparator sequence, we have  $g^+_t(u_t) = 0$ for $ 1\leq t\leq T$. Hence, the previous inequality yields the following:

 %So we can subtract the weighted sum of these terms to the right of the above equation to get 
%\cmt{done up to this}
\begin{eqnarray*}
&&\Phi\bigl(Q(T)\bigr) + V\sum\limits_{t=1}^T \Bigr(f_t(x_t) - f_t(u_t)\Bigl) \\
&\leq& \sum\limits_{t=1}^T \Bigr(Vf_t(x_t) + \Phi'(Q(t))g_t^+(x_t) \Bigr) - \sum\limits_{t=1}^T \Bigr(Vf_t(u_t) + \Phi'(Q(t))g_t^+(u_t)\Bigl)\\
&=& \sum\limits_{t=1}^T  \Bigr(\hat{f}_t(x_t) - \hat{f}_t(u_t)\Bigl),
\end{eqnarray*}
where the surrogate functions $\{\hat{f}_t\}_{t \geq 1}$ have been defined in Eqn.\ \eqref{surr-cost23}.  
Hence, using the definition of universal dynamic regret \eqref{ud-reg-def}, we obtain the following \emph{dynamic regret decomposition inequality}:
\begin{equation} \label{regretdecompositioninequality}
    \Phi(Q(T)) + V~\textsc{UD-Regret}(f_{1:T}; u_{1:T}) \leq \textsc{UD-Regret}(\hat{f}_{1:T}; u_{1:T}), 
\end{equation}
%where $\textsc{UD-Regret}_T$ and $\textsc{UD-Regret}_T'$ refer to the universal dynamic regret for the original cost functions $\{f_t\}_{t\geq 1}$ and surrogate cost functions $\{\hat{f}_t\}_{t \geq 1}$, respectively. 
Inequality \eqref{regretdecompositioninequality} plays a key role in the analysis by relating the cumulative constraint violation with the universal dynamic regret through the Lyapunov function.

Since the surrogate cost function $\hat{f}_t$, defined in Eqn.\ \eqref{surr-cost23}, involves a time-varying factor $\Phi'(Q(t)),$ its Lipschitz constant cannot be bounded \emph{a priori} as the CCV $Q(t)$ could potentially grow indefinitely. Because of this, Algorithm \ref{COCOAlgonew1} passes the surrogate costs to the Lipschitz-adaptive $\mathsf{AHAG}$ algorithm, described in Algorithm \ref{alg:AdaHedgeGrad}, which does not require any fixed upper bound to the Lipschitz constants of the cost functions. Plugging in the universal dynamic regret bound from Theorem  \ref{meta_dyn_reg_ogd2} for the surrogate cost functions, the regret decomposition inequality \eqref{regretdecompositioninequality} yields the following:
\begin{equation} \label{eq:intermediate1}
\begin{aligned}
  \Phi(Q(T)) + V~\textsc{UD-Regret}(f_{1:T}; u_{1:T}) 
 \leq \gamma \sqrt{1+\mathcal{P}_T(u_{1:T})}
    \sqrt{\sum\limits_{t=1}^T ||\nabla_t||^2},
\end{aligned}
\end{equation}
where $\nabla_t=\nabla\hat{f}_t(x_t) = V\nabla f_t(x_t) + \Phi'(Q(t))\nabla g^+_t(x_t)$ is a (sub)-gradient of the surrogate cost function at $x_t$. Furthermore, the term appearing on the RHS can be upper-bounded as follows:
\begin{eqnarray*}
\sqrt{\sum\limits_{t=1}^T ||\nabla_{t}||^2} 
    &\stackrel{(a)}{\leq}& \sqrt{\sum\limits_{t=1}^T G^2~(V + \Phi'(Q(t)))^2}\\
    &\stackrel{(b)}{\leq}& G\sqrt{\sum\limits_{t=1}^T 2V^2 + \sum_{t=1}^T2(\Phi'(Q(t)))^2}\\
    &\stackrel{(c)}{\leq}& \sqrt{2}GV\sqrt{T} + \sqrt{2}G\sqrt{\sum\limits_{t=1}^T(\Phi'(Q(t)))^2} \\
    &\stackrel{(d)}{\leq}& \sqrt{2}GV\sqrt{T} + \sqrt{2}G\Phi'(Q(T))\sqrt{T},
\end{eqnarray*}

where inequality $(a)$ follows from the $G$-Lipschitzness of the cost and constraint functions and using triangle inequality, $(b)$ follows from the fact that $(x+y)^2\leq 2x^2 +2y^2$, $(c)$ follows from the fact that for any two non-negative real numbers $x,y,$ we have $\sqrt{x+y}\leq\sqrt{x}+\sqrt{y},$ and $(d)$ follows from the monotonicity of the $Q(t)$ variables and the convexity of the $\Phi(\cdot)$ function. Plugging in the above bound in \eqref{eq:intermediate1}, we obtain 
\begin{eqnarray}\label{parameter_free_eq1}
 \Phi(Q(T)) + V~\textrm{UD-Regret}(f_{1:T};u_{1:T})  &\leq&  \gamma\sqrt{1+\mathcal{P}_T(u_{1:T})}  \Phi'(Q(T))\sqrt{T} \nonumber\\
 &&+
   \gamma\sqrt{1+\mathcal{P}_T(u_{1:T})} V\sqrt{T},
\end{eqnarray}
where we have redefined the constant $\gamma \gets \gamma G \sqrt{2}.$
Finally, choosing $\Phi(x)$ to be the quadratic potential function, $\Phi(x)=x^2,$ Eqn.\ \eqref{parameter_free_eq1} simplifies to
\begin{equation}\label{param-working-eqn}
%\begin{aligned}
    Q^2(T) + V~\textsc{UD-Regret}(f_{1:T}; u_{1:T}) \leq 2\gamma\sqrt{1+\mathcal{P}_T(u_{1:T})}\,Q(T)\sqrt{T} + \gamma\sqrt{1+\mathcal{P}_T(u_{1:T})}V\sqrt{T}.
%\end{aligned}
\end{equation}
We now solve inequality \eqref{param-working-eqn} for obtaining bounds for CCV ($Q(T)$) and $\textsc{UD-Regret}(f_{1:T}; u_{1:T}).$ 
\paragraph{Bounding CCV:} Since the cost functions are $G$-Lipschitz, we have
\begin{eqnarray*}
	|f_t(x_t)-f_t(u_t)| \leq G||x_t - u_t|| \leq GD, ~\forall t \implies \textsc{UD-Regret}(f_{1:T}; u_{1:T}) \geq -GDT. 
\end{eqnarray*}
%Using the $-GDt$ lower bound for dynamic using assumption \ref{ass-1} and rearranging, we get a quadratic inequality
Plugging in the above bound into inequality \ \eqref{param-working-eqn}, we obtain 
\begin{equation*}
%\begin{aligned}
Q^2(T) - 2\gamma\sqrt{T\big(1+\mathcal{P}_T(u_{1:T})\big)}Q(T) - \gamma V\sqrt{T\big(1+\mathcal{P}_T(u_{1:T})\big)} - VGDT \leq 0.
%\end{aligned}
\end{equation*}
Solving the above quadratic inequality in $Q(T)$, we obtain:
\begin{eqnarray*}
%\begin{aligned}
    2Q(T)&\leq& 2\gamma\sqrt{T\big(1+\mathcal{P}_T(u_{1:T})\big)}+\bigg(\big(2\gamma\sqrt{T(1+\mathcal{P}_T(u_{1:T})}\big)^2 \nonumber \\
    &+& 4\gamma V\sqrt{T\big(1+\mathcal{P}_T(u_{1:T})\big)} + 4VGDT\bigg)^\frac{1}{2}.
%\end{aligned}
\end{eqnarray*}
Using the fact that for $x\geq 0,y\geq 0,$ we have $\sqrt{x+y}\leq\sqrt{x}+\sqrt{y},$ the above inequality yields the following CCV bound:
\begin{equation}\label{Q-before-tuning}
%\begin{aligned}
Q(T)
    \leq~ 2\gamma\sqrt{T(1+\mathcal{P}_T(u_{1:T}))}
    +\frac{1}{2}(4\gamma V\sqrt{T(1+\mathcal{P}_T(u_{1:T}))})^\frac{1}{2} 
    +\frac{1}{2}(4VGDT)^\frac{1}{2}.
%\end{aligned}
\end{equation}
Since the above CCV bound holds for any feasible comparator sequence, the smallest upper bound is obtained by choosing a feasible comparator  $u^\star_{1:T}$ having the shortest path length $\mathds{P}^\star_T.$

\paragraph{Bounding Regret:} From  inequality \ \eqref{param-working-eqn}, we obtain the following bound for the universal dynamic regret
\begin{equation}\label{DReg1}
\begin{aligned}
    V~\text{UD-Regret}(f_{1:T}; u_{1:T}) &\leq 2\gamma\sqrt{T\big(1+\mathcal{P}_T(u_{1:T})\big)}Q(T) - Q^2(T)  + \gamma V\sqrt{T\big(1+\mathcal{P}_T(u_{1:T})\big)} \\
    &\stackrel{(a)}{\leq} \gamma^2(1 + \mathcal{P}_T(u_{1:T}))T +\gamma V\sqrt{T\big(1+\mathcal{P}_T(u_{1:T})\big)},
\end{aligned}
\end{equation}
where $(a)$ follows from the fact that $2aQ(T)-Q^2(T) \leq a^2, \forall Q(T) \in \mathbb{R}.$
Finally, choosing the parameter $V=\gamma T^\frac{1}{2},$  we arrive at our main result of this section.

%\cmt{Remove $\gamma$ and $N$ from the results.} 
\begin{theorem}[Universal Dynamic Regret and CCV for Algorithm \ref{COCOAlgonew1}] \label{second-alg-bd}
    %For the COCO problem with adversarial constraints and hard constraint violation, under assumptions \ref{ass-1},\ref{x-t feasibility} and \ref{ass-4}, using Algorithm \ref{COCOAlgo2} we get the following bound for universal dynamic regret against any sequence of comparators $u_{t:T}$ and CCV:
    Algorithm \ref{COCOAlgonew1} achieves the following universal dynamic regret and CCV bounds:
    \begin{eqnarray*}
   %&& %\textsc{UD-Regret}_T(u_{1:T}) \leq 5\sqrt{2}G(D+1)\sqrt{4 +\ln N}(1+\mathcal{P}_T(u_{1:T}))\sqrt{T}, \\
   && \textsc{UD-Regret}(f_{1:T}; u_{1:T}) \leq \zeta GD \big(1+\mathcal{P}_T(u_{1:T})\big)\sqrt{T}, 
 % && \textrm{CCV}_T \leq 3(1+\sqrt{GD})(5\sqrt{2}G(D+1)\sqrt{4 +\ln N})^{\frac{3}{2}}\sqrt{1+\mathds{P}^\star_T}T^\frac{3}{4},
 ~\textsc{CCV}_T \leq \zeta GD(T^{\nicefrac{3}{4}} + \sqrt{T(1+\mathds{P}_T^\star)}),
    \end{eqnarray*}
    where $u_{1:T}$ is any feasible comparator sequence, $\mathds{P}^\star_T$ is the minimum path length among all feasible comparators, \emph{i.e.,} \[\mathds{P}_T^\star = \min_{y_1,\dots,y_T} \sum\limits_{t=2}^T ||y_t - y_{t-1}||_2, \quad y_t \in \mathcal{X}_t^\star, ~~ t\geq 1,\]
    and $\zeta = O(\log \log T).$
    % and 
     %      $N = \frac{1}{2}\log_2(\lceil 1 + DT\rceil)+1$.
\end{theorem}
Theorem \ref{second-alg-bd} upper bounds \textrm{CCV} in terms of the shortest possible feasible path length $\mathbb{P}_T^\star,$ which captures the intrinsic difficulty of satisfying the time-varying constraints.

\begin{remark}
    If the common feasibility assumption (Assumption \ref{com-feas}) holds, we immediately have  $\mathds{P}^\star_T=0.$ This yields 
    $\textsc{UD-Regret}=O((1+\mathcal{P}_T(u_{1:T}))\sqrt{T}), ~~\textrm{CCV}_T = \tilde{O}(T^{\nicefrac{3}{4}}).
    $
   %Furthermore, choosing the comparator sequence $u_t=x^\star \forall t,$ where $x^\star$ is a common feasible action, we obtain the optimal $O(\sqrt{T})$ static regret bound. 
These bounds generalize prior results of \citet[Theorem 3]{guo2022online} and \citet[Section III]{cao2018online}. Although both the previous algorithms need to know the full constraint functions, Algorithm \ref{COCOAlgonew1} is computationally efficient as it is projection-free and only requires first-order feedback. 

\end{remark}
\paragraph{Comparison between Algorithm \ref{COCOAlgonewest1} and Algorithm \ref{COCOAlgonew1}:} Comparing the statements of Theorem \ref{first-alg-bd} and Theorem \ref{second-alg-bd}, we observe that the first algorithm offers an improved dependence on the path length in the universal dynamic regret bound ($\tilde{O}(\sqrt{1+\mathcal{P}_T(u_{1:T}}))$ vs $\tilde{O}((1+\mathcal{P}_T(u_{1:T}))$). On the other hand, 
%the comparison between their constraint violation (CCV) guarantees is more complex. Consider an instance where the cost functions vary more rapidly than the constraint functions. Specifically, let $\mathds{P}_T^\star \ll \mathcal{P}_T^\star,$ where we recall that $\mathcal{P}_T^\star$ corresponds to the path length generated by the minimizers of the cost functions and $\mathds{P}_T^\star$ corresponds to the minimum feasible path length. 
for problems where the cost functions change rapidly with time, the second algorithm yields a better CCV bound. To see this, recall that $\mathcal{P}_T^\star$ denotes the path length generated by the minimizers of the cost functions and $\mathds{P}_T^\star$ is the shortest feasible path length, which is independent of the cost functions. Now, if $\mathds{P}_T^\star = \Omega(\sqrt{T}),$ the second algorithm yields $O(\sqrt{T(1+\mathds{P}_T^\star)})$ CCV which could be substantially smaller than the $O(\sqrt{T(1+\mathcal{P}_T^\star)})$ CCV bound guaranteed by Algorithm \ref{COCOAlgonewest1}. 
%In this case, the second algorithm offers a better CCV bound than the former. 
On the other hand, in terms of computational complexity, the Algorithm \ref{COCOAlgonew1} is more efficient than Algorithm \ref{COCOAlgonewest1} as the former needs only the gradient information and avoids the costly projection step.  
\iffalse
\subsection{Combined Algorithm}

We now present an algorithm which combines these two algorithms to get an algorithm which under the common feasibility in the worst case gives us the bounds $\mathcal{O}\bigl((1+\mathcal{P}_T)\sqrt{T}\bigr)$ regret and $T^{\frac{3}{4}}$ CCV. This happens when the comparator sequence of loss function minimizers, $\mathcal{P_T^*}$ are of the order $\Omega(\sqrt{T})$.

\begin{algorithm}
\caption{Combined COCO Algorithm} \label{COCOAlgonew3}
\begin{algorithmic}[1]
\STATE Run COCO Algorithm \ref{COCOAlgonew1}
\STATE If $CCV > T^{\frac{3}{4}}$, then set $Q(t) = 0$ and run Algorithm \ref{COCOAlgonew2}
\end{algorithmic}
\end{algorithm}

Clearly the CCV of this Algorithm is at most $Q(T) + T^{\frac{3}{4}} + GD$ which by Theorem \ref{Thorem:COCO2} is $\mathcal{O}(T^{\frac{3}{4}})$. The dynamic regret of this algorithm is at most the sum of the dynamic regrets of the COCO Algorithms \ref{COCOAlgonew1} and \ref{COCOAlgonew2}. Their corresponding dynamic regret bounds given in Theorem \ref{Theorem:COCO1} and Theorem \ref{Thorem:COCO2}, tells us immediately without further analysis that the dynamic regret of the combined COCO algorithm given above as $\mathcal{O}\bigl(\sqrt{(1+\mathcal{P_T})T}\bigr) +\mathcal{O}\bigl((1+\mathcal{P_T})\sqrt{T}\bigr) = \mathcal{O}\bigl((1+\mathcal{P_T})\sqrt{T}\bigr)$.
\fi

%\input{algo}
%\input{bandit}
% \input{Parameter-free}
% \input{NewResults}
% \input{AlternativeSurrogate}
% \input{experiments}
\section{Conclusion} \label{conclusion}
In this paper, we propose two different online algorithms for COCO, each achieving sublinear universal dynamic regret and cumulative constraint violation bounds. These bounds improve upon state-of-the-art results and remove the restrictive common feasibility assumption made in the literature. Our proposed algorithms have a modular structure and come with a simple and streamlined analysis using regret decomposition inequality. An important open direction is to establish simultaneous lower bounds on the universal dynamic regret and cumulative constraint violation. It will also be interesting to extend the proposed framework to bandit feedback settings.

%In future, it will be interesting to establish lower bounds that hold simultaneously for the universal regret and CCV and extend the algorithms to the bandit feedback settings.   
\section{Acknowledgment} \label{ack}
We thank the anonymous reviewers for their constructive feedback, which helped improve the clarity of the presentation. This work was supported in part by the Department of Atomic Energy, Government of India, under project no.\ RTI4001 and in part by a Google India Faculty Research Award. 
\clearpage
\bibliography{OCO.bib}
\clearpage
\onecolumn
\section{Appendix} \label{appendix}

\subsection{Visualizing the transformed functions} \label{f-tilde-fig}
\paragraph{Function of one variable:}
\begin{figure}[!h]
\centering
	\includegraphics[scale=1]{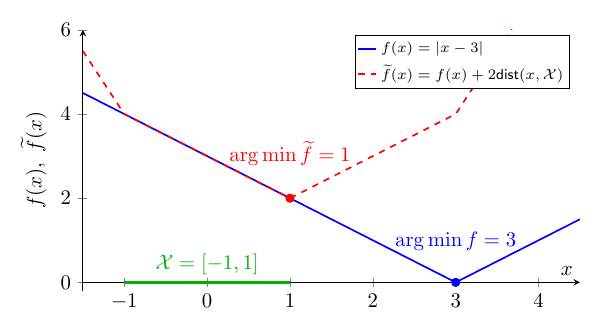}
	\caption{\small{Illustration of the transformation $f  \rightsquigarrow \widetilde f.$ 
Here $f(x)=|x-3|$, $\mathcal{X}=[-1,1], G=1$, and $\widetilde f(x)=f(x)+2\,\mathsf{dist}(x,\mathcal{X})$. 
After the transformation, the original minimizer at $x=3$ is sent to $x=1$, which lies at the boundary of $\mathcal{X}$.}}
	\label{1D-fun}
\end{figure}
Figure \ref{1D-fun} below illustrates a single variable convex function $f(x)=|x-3|, x \in \mathbb{R}$ and the associated auxiliary cost function $\tilde{f}(x).$ The feasible set is chosen to be the closed interval $\mathcal{X}=[-1,1].$ From the figure, we observe that while the unconstrained minimum value of the function $f$ is attained at $x=3,$ the unconstrained minimum of the auxiliary function $\tilde{f}$ is attained at $x=1,$ which is \emph{in} the feasible set. This verifies the conclusion from Lemma \ref{lemma:9.2}.
 
\begin{figure}[H]
\centering
	\includegraphics[scale=0.7]{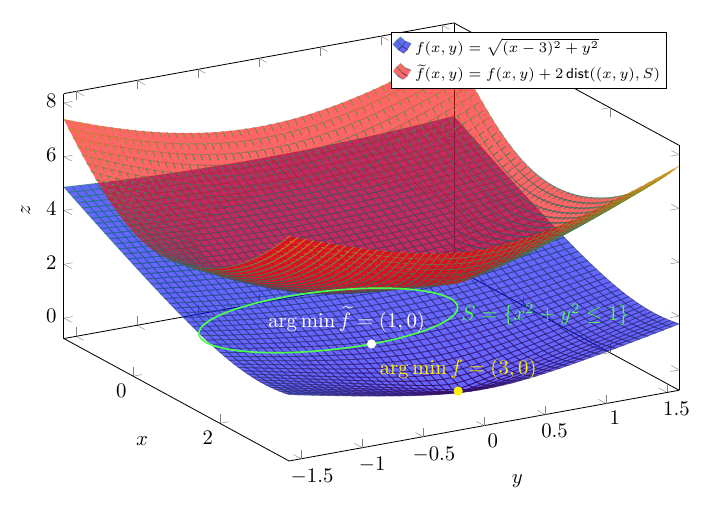}
	\caption{\small{Illustration of the transformation $f \rightsquigarrow \widetilde f.$ 
Here $f(x,y)=\sqrt{(x-3)^2+y^2}, G=1,$ feasible set $S=\{(x,y):x^2+y^2\le1\}$, and $\widetilde f(x,y)=f(x,y)+2\,\mathrm{dist}((x,y),S)$. 
After the transformation, the original minimizer at $(3,0)$ is sent to $(1,0)$ which lies at the boundary of $S$.}}
	\label{2D-fun}
\end{figure}

\paragraph{Function of two variables:} Figure \ref{2D-fun} illustrates a similar observation for a convex function with two variables defined as $f(x) = \sqrt{(x-3)^2+y^2}, ~x, y \in \mathbb{R},$ with the convex feasible set $S=\{(x,y) \in \mathbb{R}^2: x^2 + y^2 \leq 1\}.$ The global minimum of the auxiliary function is shifted to the feasible point $(1,0)$ from the global minimum of the original function located at $(3,0).$

\subsection{Necessity of the $\dist(\cdot, \mathcal{X}_t^\star)$ term in the definition of the surrogate function \eqref{surr-cost-def-alg2}} \label{dist-nec}
\begin{figure*}
	\centering
	\includegraphics[scale=0.8]{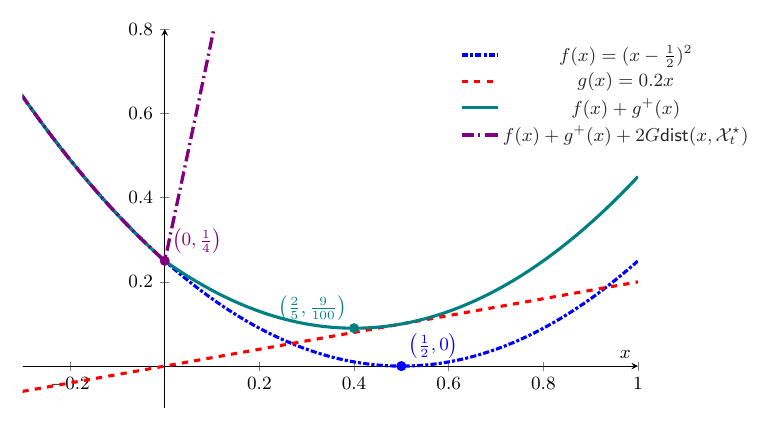}
	\caption{\small{Illustrating various original and the surrogate cost functions and their minima}}
	\label{dist_func}
\end{figure*}

In this section, we give a simple counterexample that shows that running \emph{any} no-regret OCO algorithm on the surrogate cost functions $\hat{f}_t := f_t+g_t^+, t \geq 1,$ may lead to a \emph{linear} CCV. Consider the following one-dimensional time-invariant problem where the decision set is $\mathcal{X} = [-1, 1]$ and the cost and constraint functions are given by  $$f_t(x) = \big(x-\frac{1}{2}\big)^2, ~~ \textrm{and}~~ g_t(x)= 0.2x, ~~ t \geq 1.$$  Clearly, all feasible sets are identical and given by $\mathcal{X}_t^\star = [-1,0], ~t \geq 1.$ The minimizer $x^\star$ of the surrogate cost function can be found by setting the derivative of $\hat{f}_t(x) \equiv \big(x-\frac{1}{2}\big)^2 + 0.2x^+ $ to zero, yielding $x^\star=\nicefrac{2}{5}.$ This implies that the iterates of \emph{any} no-regret policy must converge to the point $x^\star =\nicefrac{2}{5}$ as $T$ gets large. Hence, after $T$ rounds, the CCV is approximately $\approx 0.2 \times \frac{2}{5} \times T = 0.08T,$ which grows \emph{linearly} with $T.$ See Figure \ref{dist_func} for an illustration. 

 The reason behind the above issue is that the magnitude of the gradient of the constraint function is negligible compared to that of the cost function. Algorithm 2 avoids this issue by adding the $\mathsf{dist}(\cdot)$ function to the surrogate cost. In this case, the surrogate cost becomes $$\hat{f}_t(x) \equiv \big(x-\frac{1}{2}\big)^2 + 0.2x^+ + 6x^+, ~~\forall t\geq 1. $$ 
It can be verified that the unconstrained minimum of the above surrogate cost function lies at $x^\star=0 \in \mathcal{X}_t^\star$ (see Lemma \ref{lemma:9.2}), leading to a sublinear CCV bound. 
% See also our response to the comments of Reviewer RTNW below.

\subsection{Proof of Theorem \ref{dyn_reg_ogd}} \label{dyn_reg_ogd_proof}
Define the sequence $y_{t+1} \vcentcolon=  x_t - \eta_t \nabla_t, ~\forall t \geq 1,$ where $\{\eta_t\}_{t \geq 1}$ is any non-increasing adaptive step size sequence. From Algorithm \ref{ogd_alg}, we have $x_{t+1}= \mathsf{Proj}_{\mathcal{X}}(y_{t+1}),~ \forall t\geq 1.$ Let $x^\star_{1:T}$ be any comparator sequence with $x^\star_t \in \mathcal{X}, \forall t \in [T].$ Then we have
\begin{eqnarray*}
||x_{t+1} - x^\star_t||^2 
\stackrel{(a)}{\leq} ||y_{t+1} - x^\star_t||^2 
= ||x_t - x^\star_t||^2 + \eta_{t}^2 ||\nabla_t||^2 - 2 \eta_t \nabla_{t}^\top (x_t - x^\star_t),
\end{eqnarray*}
where inequality (a) follows from the non-expansive property of Euclidean projection and the second equality follows from the definition of $y_{t+1}$.
Rearranging the above inequality, we have:
\begin{equation}
    2 \nabla_{t}^\top(x_t - x^\star_t) \leq \frac{||x_t - x^\star_t||^2 - ||x_{t+1} - x^\star_t||^2 }{\eta_t} + \eta_t ||\nabla_t||^2.
\end{equation}

Using the convexity of the cost functions and summing the above inequalities for $1\leq t \leq T$, we obtain:
\begin{eqnarray} \label{dyn-reg-eq}
2 \sum_{t=1}^{T} \left( \hat{f}_t(x_t) - \hat{f}_t(x^\star_t) \right) 
\leq 2\sum_{t=1}^T \nabla_{t}^\top(x_t - x^\star_t) 
\leq \underbrace{\sum_{t=1}^T \frac{||x_t - x^\star_t||^2 - ||x_{t+1} - x^\star_t||^2 }{\eta_t}}_{(A)} + \sum_{t=1}^T \eta_t ||\nabla_t||^2. 
\end{eqnarray}
%\cmt{Since $\eta_t$ is monotone, the result should be immediate. See Hazan.}
%The RHS of the expression in \eqref{dyn-reg-eq} can be rewritten as
Next we simplify term (A) in Eqn. \eqref{dyn-reg-eq}. We have
\begin{flalign} \label{termB}
   (A) & = \frac{||x^\star_1 - x_1||^2}{\eta_1} - \frac{||x^\star_T - x_{T+1}||^2}{\eta_T} + \sum\limits_{t=1}^{T-1} \frac{||x_{t+1} - x^\star_{t+1}||^2}{\eta_{t+1}} - \frac{||x_{t+1} - x^\star_t||^2}{\eta_t}  \nonumber\\
    & =\frac{||x^\star_1 - x_1||^2}{\eta_1} - \frac{||x^\star_T - x_{T+1}||^2}{\eta_T} +
     \underbrace{\sum\limits_{t=1}^{T-1} \frac{\eta_t||x_{t+1} - x^\star_{t+1}||^2 - \eta_{t+1}||x_{t+1} - x^\star_t||^2}{\eta_t\eta_{t+1}}}_\text{\clap{(B)}}.
    % &\leq \sum\limits_{t=1}^T \frac{||x_t + x_{t+1} - 2x^\star_t||\quad||x_t - x_{t+1}||}{\eta_t} + \sum\limits_{t=1}^T \eta_t ||\nabla_t||^2
    % && \parbox[t]{0.5\textwidth}{\raggedleft (Using Cauchy-Schwartz)} \\
    % &\leq 2D\sum\limits_{t=1}^T \frac{||x_t - x_{t+1}||}{\eta_t} + \sum\limits_{t=1}^T \eta_t ||\nabla_t||^2
    % && \parbox[t]{0.5\textwidth}{\raggedleft (Using triangle inequality and upper bounding by the diameter of the feasible set)} \\
\end{flalign}
We next upper bound term (B) in \eqref{termB}. We have 
\begin{flalign*}
    (B) &=\sum\limits_{t=1}^{T-1} \frac{||\sqrt{\eta_t}x_{t+1} - \sqrt{\eta_t}x^\star_{t+1}||^2 - ||\sqrt{\eta_{t+1}}x_{t+1} - \sqrt{\eta_{t+1}}x^\star_t||^2}{\eta_t\eta_{t+1}} \\
    &= \sum\limits_{t=1}^{T-1} \frac{\langle (\sqrt{\eta_t} + \sqrt{\eta_{t+1}})x_{t+1} - \sqrt{\eta_t}x^\star_{t+1} - \sqrt{\eta_{t+1}}x^\star_t, (\sqrt{\eta_t} - \sqrt{\eta_{t+1}})x_{t+1} - \sqrt{\eta_t}x^\star_{t+1} + \sqrt{\eta_{t+1}}x^\star_t \rangle}{\eta_t\eta_{t+1}} \\
    &\leq \sum\limits_{t=1}^{T-1} \frac{||(\sqrt{\eta_t} + \sqrt{\eta_{t+1}})x_{t+1} - \sqrt{\eta_t}x^\star_{t+1} - \sqrt{\eta_{t+1}}x^\star_t|| \: ||(\sqrt{\eta_t} - \sqrt{\eta_{t+1}})x_{t+1} - \sqrt{\eta_t}x^\star_{t+1} + \sqrt{\eta_{t+1}}x^\star_t||}{\eta_t\eta_{t+1}},   
\end{flalign*}
where the last step follows from Cauchy-Schwarz inequality. Note that the first term in the numerator can be bounded as:
%\begin{flalign*}
  \[  ||(\sqrt{\eta_t} + \sqrt{\eta_{t+1}})x_{t+1} - \sqrt{\eta_t}x^\star_{t+1} - \sqrt{\eta_{t+1}}x^\star_t|| 
    \leq \sqrt{\eta_t}||x_{t+1} - x^\star_{t+1}|| + \sqrt{\eta_{t+1}} ||x_{t+1} - x^\star_t||
    \leq (\sqrt{\eta_t} + \sqrt{\eta_{t+1}}) D ,\]
%\end{flalign*}
where we have used the triangle inequality. 
Using this, we have
\begin{flalign*}
 (B)   &\leq D\sum\limits_{t=1}^{T-1} \frac{ (\sqrt{\eta_t} + \sqrt{\eta_{t+1}}) \: ||(\sqrt{\eta_t} - \sqrt{\eta_{t+1}})x_{t+1} - \sqrt{\eta_t}x^\star_{t+1} + \sqrt{\eta_{t+1}}x^\star_t||}{\eta_t\eta_{t+1}}\\
    &= D\sum\limits_{t=1}^{T-1} \frac{ (\sqrt{\eta_t} + \sqrt{\eta_{t+1}}) ||(\sqrt{\eta_t} - \sqrt{\eta_{t+1}})(x_{t+1}-x^\star_{t+1}) + \sqrt{\eta_{t+1}}(x^\star_t - x^\star_{t+1})||}{\eta_t\eta_{t+1}}\\
    %[(\sqrt{\eta_t} - \sqrt{\eta_{t+1}})||x_{t+1}|| + ||\sqrt{\eta_{t+1}}x^\star_t - \sqrt{\eta_t}x^\star_{t+1}||]}{\eta_t\eta_{t+1}} \\
%    && \parbox[t]{0.2\textwidth}{\raggedleft (Using triangle inequality)} \\
    &\stackrel{(a)}{\leq} D\sum\limits_{t=1}^{T-1} \frac{ (\sqrt{\eta_t} + \sqrt{\eta_{t+1}})  \: 
    [(\sqrt{\eta_t} - \sqrt{\eta_{t+1}})D + \sqrt{\eta_{t+1}}||x^\star_t - x^\star_{t+1}||]}{\eta_t\eta_{t+1}}\\
%\end{flalign*}
 %Now, 
%
%\begin{flalign*}
%    &||\sqrt{\eta_{t+1}}x^\star_t - \sqrt{\eta_t}x^\star_{t+1}|| \\
%    &=||\sqrt{\eta_{t+1}}x^\star_t - \sqrt{\eta_{t+1}}x^\star_{t+1} + \sqrt{\eta_{t+1}}x^\star_{t+1} - \sqrt{\eta_t}x^\star_{t+1}|| \\
%    &\leq \sqrt{\eta_{t+1}}||x^\star_{t+1} - x^\star_{t}|| + (\sqrt{\eta_{t+1}} - \sqrt{\eta_{t}})||x^\star_{t+1}|| \\ 
%    &\leq \sqrt{\eta_{t+1}}||x^\star_{t+1} - x^\star_{t}|| + (\sqrt{\eta_{t}} - \sqrt{\eta_{t+1}})\frac{D}{2}
%\end{flalign*}
%Using this, term (a)
%\begin{flalign*}
   % &\leq \sum\limits_{t=1}^{T-1} \frac{ (\sqrt{\eta_t} + \sqrt{\eta_{t+1}}) D \: 
    %[(\sqrt{\eta_t} - \sqrt{\eta_{t+1}})\frac{D}{2} + (\sqrt{\eta_{t}} - \sqrt{\eta_{t+1}})\frac{D}{2} + \sqrt{\eta_{t+1}}||x^\star_{t+1} - x^\star_{t}||]}{\eta_t\eta_{t+1}} \\
    %&=\sum\limits_{t=1}^{T-1} \frac{ (\sqrt{\eta_t} + \sqrt{\eta_{t+1}}) D \: 
    %[(\sqrt{\eta_t} - \sqrt{\eta_{t+1}})D + \sqrt{\eta_{t+1}}||x^\star_{t+1} - x^\star_{t}||]}{\eta_t\eta_{t+1}} \\
    &\stackrel{(b)}{\leq} D\sum\limits_{t=1}^{T-1} \frac{(\eta_t - \eta_{t+1}) D \: 
    + 2\eta_t ||x^\star_{t+1} - x^\star_{t}||}{\eta_t\eta_{t+1}} \\
    %&\leq \sum\limits_{t=1}^{T-1} \frac{\eta_t D^2 \: 
    %+ 2\eta_t D||x^\star_{t+1} - x^\star_{t}||}{\eta_t\eta_{t+1}} \\
    &= D^2 \big(\frac{1}{\eta_T} - \frac{1}{\eta_1}\big) + 2D \sum\limits_{t=1}^{T-1} \frac{||x^\star_{t+1} - x^\star_{t}||}{\eta_{t+1}}\\
    &\stackrel{(c)}{\leq}  D^2 \big(\frac{1}{\eta_T} - \frac{1}{\eta_1}\big) + \frac{2D \mathcal{P}_T(x^\star_{1:T})}{\eta_T},
    % &= \sum\limits_{t=1}^{T-1} \frac{ (\sqrt{\eta_t} + \sqrt{\eta_{t+1}}) D \:
    % \sqrt{\eta_{t+1}}||x^\star_{t+1} - x^\star_{t}||}
    % {\eta_t\eta_{t+1}}\\
    % &\leq \sum\limits_{t=1}^{T-1} \frac{ 2\eta_t D \: ||x^\star_{t+1} - x^\star_{t}||}
    % {\eta_t\eta_{t+1}} 
    % && \parbox[t]{0.1\textwidth}{\raggedleft (Using \(\eta_{t+1} \leq \eta_{t})\)} \\
    % &\leq 2 D \sum\limits_{t=1}^{T-1} \frac{||x^\star_{t+1} - x^\star_{t}||}
    % {\eta_{t+1}} \\
    % &\leq \frac{2 D}{\eta_{T}} \sum\limits_{t=1}^{T-1} ||x^\star_{t+1} - x^\star_{t}|| \\
    % &= \frac{2 D \mathcal{P}^T}{\eta_{T}}
\end{flalign*}
where in step (a), we have used the triangle inequality, and in step (c), we have used the definition of the path length of the comparator. The non-increasing property of the step sizes, \emph{i.e.,} $\eta_t \geq \eta_{t+1}, \forall t \geq 1$ was used in steps (a), (b), and (c). Finally, combining the above bound with Eqns. \eqref{dyn-reg-eq} and \eqref{termB}, we conclude 
\begin{flalign*}
    & 2(\sum\limits_{t=1}^{T} \hat{f}_t(x_t) - \hat{f}_t(x^\star_t)) \\
    &\leq \frac{||x^\star_1 - x_1||^2}{\eta_1} - \frac{||x^\star_T - x_{T+1}||^2}{\eta_T} + D^2 \big(\frac{1}{\eta_T} - \frac{1}{\eta_1}\big)
      + \frac{2 D \mathcal{P}_T(x^\star_{1:T})}{\eta_{T}}
      + \sum\limits_{t=1}^T \eta_t ||\nabla_t||^2\\
   % &\leq \frac{D^2}{\eta_T}
    %  + \frac{2 D \mathcal{P}^T}{\eta_{T}}
    %  + \sum\limits_{t=1}^T \eta_t ||\nabla_t||^2\\
    &\leq \frac{D^2 + 2 D \mathcal{P}_T(x^\star_{1:T})}{\eta_T}
      + \sum\limits_{t=1}^T \eta_t ||\nabla_t||^2.
\end{flalign*}
Hence, the universal dynamic regret \eqref{ud-reg-def} of Algorithm \ref{ogd_alg} can be upper bounded as 
\begin{equation}\label{eqn2}
    \textsc{UD-Regret}(\hat{f}_{1:T}; x^\star_{1:T}) \leq \frac{\max(D^2,2D)(1 + \mathcal{P}_T(x^\star_{1:T}))}{2\eta_T}
      + \underbrace{\frac{1}{2}\sum\limits_{t=1}^T \eta_t ||\nabla_t||^2}_{(C)}.
\end{equation}
The dynamic regret bound in Eqn.\ \eqref{eqn2} holds for Algorithm \ref{ogd_alg} with any non-increasing step sizes. Assuming the path-length is known to be bounded as $\mathcal{P}_T(x^\star_{1:T}) \leq \mathcal{P}_T,$ using the specific choice of the step size sequence $\eta_t = \frac{(D+1) \sqrt{1+\mathcal{P}_T}}{\sqrt{2\sum\limits_{\tau = 1}^t ||\nabla_\tau||^2}}, t \geq 1,$  we can upper bound term (C) as follows 
  \begin{flalign*}
\frac{1}{2}\sum\limits_{t=1}^T \eta_t ||\nabla_t||^2 &=
\frac{(D+1)\sqrt{1+\mathcal{P}_T}}{2\sqrt{2}}  \sum_{t=1}^T\frac{||\nabla_t||^2}{\sqrt{\sum\limits_{\tau=1}^t ||\nabla_\tau||^2} } \\
&\leq \frac{(D+1)\sqrt{1+\mathcal{P}_T}}{2\sqrt{2}} \int_{0}^{\sum\limits_{t=1}^T ||\nabla_t||^2} \frac{dz}{\sqrt{z}} \\
&= \frac{(D+1)\sqrt{1+\mathcal{P}_T}}{\sqrt{2}}\sqrt{\sum\limits_{t=1}^T ||\nabla_t||^2}.
\end{flalign*}
Hence, from \eqref{eqn2}, the dynamic regret for Algorithm \ref{ogd_alg} can be bounded as 
\begin{equation}\label{eqn3}
    \textsc{D-Regret}(\hat{f}_{1:T}; \mathcal{P}_T) \leq (D+1) \sqrt{2(1+\mathcal{P}_T)}
    \sqrt{\sum\limits_{t=1}^T ||\nabla_t||^2}.
\end{equation}
In a similar fashion, plugging in the path-length independent step size sequence $\eta_t = \frac{(D+1)}{ \sqrt{2\sum_{\tau=1}^t ||\nabla_\tau||^2}}, ~t \geq 1,$ into \eqref{eqn2}, we obtain the universal dynamic regret bound \eqref{ud-regret-bd}. 
\subsection{Proof of Theorem \ref{meta_dyn_reg_ogd2}} \label{meta_dyn_reg_ogd_proof}
%\begin{proofof}
Since $\text{diam}(\mathcal{X}) \leq D,$ the path length $\mathcal{P}_T(u_{1:T})$ of any comparator sequence can be trivially bounded as  $0 \leq \mathcal{P}_T(u_{1:T}) \leq DT.$
%\[
%0 \quad\leq \quad \sqrt{1 +\mathcal{P}_T}\quad \leq \quad \sqrt{1+DT},
%\]
%\cmt{Why $\sqrt{1+DT}?$}
 Algorithm \ref{alg:AdaHedgeGrad} maintains $N:=\lceil\frac{1}{2}\log_2(1+DT)\rceil + 1$ experts where the $i$\textsuperscript{th} expert estimates the path length to be $\rho_i := 2^{i-1},~ i \in [N]$. 
%\[
%N:=\lceil\frac{1}{2}\log_2(1+DT)\rceil + 1, ~~ 
%\rho_i := 2^{i-1}, \quad \text{for} \,\, i \in \{1, \dots, N \}
%\]
This ensures that for each candidate path length $\mathcal{P}_T(u_{1:T}),$ there is an estimate $\rho_i$ which approximates the true path length within a constant factor, \emph{i.e.,}
\begin{equation}\label{rhochoose}    
\forall \mathcal{P}_T(u_{1:T}),\,\exists \, \rho_i \quad \text{s.t.} \quad \frac{1}{2}\rho_i \leq \sqrt{1+\mathcal{P}_T(u_{1:T})} \leq \rho_i.
\end{equation}
%The idea is to track $N$ experts, with expert $i$ using $\rho_i$ as their guess for the path length. 
At round $t$, the $i^\text{th}$ expert predicts $x^i_t \in \mathcal{X}$ using $\mathsf{AdaGrad}$ (Algorithm \ref{ogd_alg}). Let us define $X_t$ to be a $d \times N$ matrix with $x_t^i$ being its $i^\text{th}$ column. The expert-tracking algorithm $\mathsf{AdaHedge}$ gives a probability distribution $w_t \in \Delta_{N-1}$. The final action at round $t$ is chosen to be a convex combination of all experts' output, \emph{i.e.,} $x_t = X_tw_t.$ 
%Also note that choosing a "pure" expert $i$'s strategy, $x_t^i$, would correspond to choosing the standard basis vector $e_i$ as the output of the meta-algorithm.

Now for a comparator sequence $u_{1:T}$, assume that Eqn.\ \eqref{rhochoose} holds for some $i \in [N].$ Then we have 
\begin{equation}
\begin{aligned}\label{eq12}
    & \textsc{UD-Regret}(f_{1:T}; u_{1:T}) \\
    & = \sum\limits_{t=1}^{T} f_t(x_t) - \sum\limits_{t=1}^{T} f_t(u_t) \\
    &\stackrel{(a)}{=} \sum\limits_{t=1}^{T} f_t(x_t) -
    \sum\limits_{t=1}^{T} f_t(x_t^i) + 
    \sum\limits_{t=1}^{T} f_t(x_t^i) -
    \sum\limits_{t=1}^{T} f_t(u_t) \\
    &\stackrel{(b)}{\leq} \underbrace{\sum\limits_{t=1}^{T} \langle  X_t^T \nabla f_t(X_tw_t)\,,\, w_t - e_i \rangle}_{\clap{\text{term (I)}}}  +
    % -
   % \sum\limits_{t=1}^{T} \langle \nabla^\intercal f_t(X_tw_t)X_t\,,\, e_i \rangle} 
   \underbrace{\sum\limits_{t=1}^{T} f_t(x_t^i) -
    \sum\limits_{t=1}^{T} f_t(u_t)}_{\clap{\text{term (II)}}},
\end{aligned}
\end{equation}
where inequality (b) follows from the convexity of the cost function $f_t$. %This is a standard trick to convert an Online Convex Optimization(OCO) problem to an Online Linear Optimization(OLO) problem. 
Next, we bound term (I) and term (II) separately. 

\paragraph{Bounding term (I):} To bound term (I), we invoke the following result from \citet[Section 7.6]{orabona2019modern} concerning the $\mathsf{AdaHedge}$ algorithm. The setup for the $\mathsf{AdaHedge}$ algorithm is the same as the classic Prediction with Experts' Advice (PEA) problem with $N$ experts \citep{cesa2006prediction}. The main distinguishing feature of $\mathsf{AdaHedge}$ compared to $\mathsf{Hedge}$ is that $\mathsf{AdaHedge}$ is adaptive to the range of the loss vectors, and hence, no uniform upper bound to the loss vectors is necessary to tune its learning rate.

\begin{theorem}[Static Regret bound of $\mathsf{AdaHedge}$] \label{adareg}
    For the Prediction with Expert Advice problem, the $\mathsf{AdaHedge}$ algorithm, when run sequentially for $T$ rounds with the loss vectors $l_1,\dots,l_T \in \mathbb{R}^N,$ achieves the following static regret bound with respect to any fixed expert $i \in [N]:$
    %with respect to any reference probability distribution $p$ on $N$ experts
    \[
    \textrm{Regret}_T(i) \leq 2\sqrt{(4+\ln{N})\sum\limits_{t=1}^{T}||l_t||_\infty^2}.
    \]
\end{theorem}

We now apply the above result to the linear loss functions $\langle l_t, w\rangle = \langle X_t^T \nabla f_t(X_tw_t), w\rangle, t \geq 1$. We can bound the $\ell_\infty$-norm of the $t$\textsuperscript{th} loss vector as follows: 
\begin{equation}\label{ell_infty_bound}
\begin{aligned}
    ||X_t^T\nabla f_t(X_tw_t)||_\infty 
    = \max_{1\leq i \leq N} |\langle x_t^i, \nabla f_t(X_tw_t) \rangle |  
    \stackrel{(a)}{\leq} ||\nabla f_t(X_tw_t)||_2 \, 
   ||x_t^i||_2
    \stackrel{(b)}{\leq}\frac{D}{2}||\nabla_t||_2, 
\end{aligned}
\end{equation}
where in (a) we have used the Cauchy–Schwarz inequality and (b) follows from the boundedness of the decision set and the assumption that the decision set lies in an Euclidean ball of radius $D$.
Finally, this yields the following upper bound for term (I):
\begin{equation}
    \textrm{Term (I)} \leq D\sqrt{4 + \ln N}\sqrt{ \sum\limits_{t=1}^{T}||\nabla_t||_2^2 }.
\end{equation}
\paragraph{Bounding term (II):} %Note that the equality (a) in \eqref{eq12} holds for any choice of $x_t^i$, hence to bound term (II), we will be using the term closest to the unknown path length. That is to say,
Recall that the expert $i$ is defined to be such that its  $\rho_i$ satisfies \eqref{rhochoose}. Furthermore, expert $i$ runs $\mathsf{AdaGrad}$ with the learning rate
\begin{equation}\label{expert_stepsize}
\eta^i_t = \frac{(D+1) \rho_i}{\sqrt{2\sum\limits_{\tau = 1}^t ||\nabla f(x_\tau^i)||^2}}, t \geq 1.   
\end{equation}
%Putting this choice of $\eta_t$ in \eqref{eqn2} in the analysis of the OGD algorithm, we get
Using Eqn.\ \eqref{d-regret-bd} from Theorem \ref{dyn_reg_ogd}, the universal dynamic regret of the $i$\textsuperscript{th} expert can be bounded as: 
\begin{equation}
\begin{aligned}
\textrm{Term (II)}\leq\textrm{D-Regret}_T(\mathcal{P}_T(u_{1:T}))\leq 2(D+1) \sqrt{2(1+\mathcal{P}_T(u_{1:T}))}
    \sqrt{\sum\limits_{t=1}^T ||\nabla_t||^2}.
\end{aligned}    
\end{equation}

Combining the bounds for terms (I) and (II), we arrive at the desired result.
\end{document}